\documentclass[manuscript,screen]{acmart}

\AtBeginDocument{%
  \providecommand\BibTeX{{%
    \normalfont B\kern-0.5em{\scshape i\kern-0.25em b}\kern-0.8em\TeX}}}

\usepackage{ragged2e}  
\usepackage{seqsplit} 
\setcopyright{acmcopyright}
\copyrightyear{2024}
\acmYear{2024}
\acmDOI{XXXXXXX.XXXXXXX}
\usepackage{enumitem}
\usepackage{hhline}  % 在导言区添加这个包  
\usepackage{hyperref}  
\usepackage{hyperxmp}  
\usepackage{longtable}  
\usepackage{array}  
\usepackage{multirow}  
\usepackage{booktabs}  
\usepackage{bm,multirow,tablefootnote}
\usepackage{longtable}
\usepackage{bbding}
\usepackage{threeparttable} 
\usepackage[para]{footmisc}
\usepackage{geometry}
\usepackage{makecell}
\usepackage{xltabular}  
\usepackage{booktabs}  
\usepackage{multirow}  
\usepackage{makecell}  
\usepackage{array}  
\usepackage{hyperref}  
\usepackage{array}  
\usepackage{ragged2e}  
\usepackage{seqsplit}  
\usepackage{url}  
\newcolumntype{P}[1]{>{\RaggedRight\arraybackslash}p{#1}}  
\usepackage{adjustbox}  
\usepackage{xcolor}
\usepackage{graphicx}  
\usepackage{array}  
\usepackage{multirow}  
\usepackage{booktabs} 
\usepackage{makecell}
\makeatletter  
\g@addto@macro{\UrlBreaks}{\UrlOrds}  
\makeatother 

\usepackage{setspace}  
\usepackage{geometry}
\begin{document}
% \begin{spacing}{1.15}
% \setstretch{1.25}
\title{Vision Technologies with Applications in Traffic Surveillance Systems: A Holistic Survey}

\author{Wei Zhou}
\email{weizhou@njust.edu.cn}
\orcid{0000-0003-3225-0576}
\affiliation{%
  \institution{School of Automation, Nanjing University of Science and Technology}
  % \streetaddress{P.O. Box 1212}
  \city{Nanjing}
  \postcode{210094}
  \country{China}
}

\author{Li Yang}
\email{yangli945@njust.edu.cn}
\orcid{0000-0002-8152-7642}
\affiliation{%
  \institution{School of Automation, Nanjing University of Science and Technology}
  % \streetaddress{P.O. Box 1212}
  \city{Nanjing}
  \postcode{210094}
  \country{China}
}
            
\author{Lei Zhao}
\email{lei\_zhao@seu.edu.cn}
\orcid{0000-0002-6336-048X}
\affiliation{%
  \institution{School of Transportation, Southeast University}
  % \streetaddress{2 Southeast University Road}
  \city{Nanjing}
  \postcode{211189}
  % \state{Jiangsu Province}
  \country{China}
}

\author{Runyu Zhang}
\email{ry01@seu.edu.cn}
\orcid{0009-0007-6278-9915}
\affiliation{%
  \institution{School of Transportation, Southeast University}
  % \streetaddress{2 Southeast University Road}
  \city{Nanjing}
  \postcode{211189}
  % \state{Jiangsu Province}
  \country{China}
}

\author{Yifan Cui}
\email{220243469@seu.edu.cn}
\orcid{0009-0002-2322-7841}
\affiliation{%
  \institution{School of Transportation, Southeast University}
  % \streetaddress{2 Southeast University Road}
  \city{Nanjing}
  \postcode{211189}
  % \state{Jiangsu Province}
  \country{China}
}

\author{Hongpu Huang}
\email{220223065@seu.edu.cn}
\orcid{0009-0003-0210-6589}
\affiliation{%
  \institution{School of Transportation, Southeast University}
  % \streetaddress{2 Southeast University Road}
  \city{Nanjing}
  \postcode{211189}
  % \state{Jiangsu Province}
  \country{China}
}

\author{Kun Qie}
\email{qiekun@stu.bucea.edu.cn}
\orcid{0009-0008-8406-7996}
\affiliation{%
  \institution{Beijing Laboratory of General Aviation Technology, Beijing University of Civil Engineering and Architecture}
  % \streetaddress{No.1 Zhanlanguan Road, Xicheng District, Beijing}
  \city{Beijing}
  \postcode{100044}
  % \state{Beijing}
  \country{China}
}

\author{Chen Wang}
\authornote{Prof. Chen Wang is the corresponding author of this paper.}
\email{chen\_david\_wang@seu.edu.cn}
\orcid{0000-0003-4573-9047}
\affiliation{%
  \institution{School of Transportation, Southeast University}
  % \streetaddress{P.O. Box 1212}
  \city{Nanjing}
  \postcode{211189}
  \country{China}
}

\renewcommand{\shortauthors}{Wei Zhou \textit{ et al.}}
\linepenalty=1000

\begin{abstract}

Traffic Surveillance Systems (TSS) have become increasingly crucial in modern intelligent transportation systems, with vision technologies playing a central role for scene perception and understanding. While existing surveys typically focus on isolated aspects of TSS, a comprehensive analytical framework bridging low-level and high-level perception tasks, particularly considering emerging technologies, remains lacking. This paper presents a systematic review of vision technologies in TSS, examining both low-level perception tasks (object detection, classification, and tracking) and high-level perception tasks (parameter estimation, anomaly detection, and behavior understanding). Specifically, we first provide a detailed methodological categorization and comprehensive performance evaluation for each task. Our investigation reveals five fundamental limitations in current TSS: perceptual data degradation in complex scenarios, data-driven learning constraints, semantic understanding gaps, sensing coverage limitations and computational resource demands. To address these challenges, we systematically analyze five categories of current approaches and potential trends: advanced perception enhancement, efficient learning paradigms, knowledge-enhanced understanding, cooperative sensing frameworks and efficient computing frameworks, critically assessing their real-world applicability. Furthermore, we evaluate the transformative potential of foundation models in TSS, which exhibit remarkable zero-shot learning abilities, strong generalization, and sophisticated reasoning capabilities across diverse tasks. This review provides a unified analytical framework bridging low-level and high-level perception tasks, systematically analyzes current limitations and solutions, and presents a structured roadmap for integrating emerging technologies, particularly foundation models, to enhance TSS capabilities.

\end{abstract}





\maketitle

\section{Introduction}

Traffic Surveillance Systems (TSS) play a crucial role in Intelligent Transportation Systems (ITS), enabling comprehensive perception and analysis of traffic scenarios. While ITS employs various sensing technologies, including inductive loops, microwaves, radar, and LiDAR, surveillance cameras have emerged as the predominant choice for traffic monitoring. This preference is primarily attributed to cameras’ unique advantages in providing continuous, high-resolution visual data with rich semantic information about traffic participants and infrastructure \cite{zhou2023pedestrian}. These distinctive capabilities have established cameras as the cornerstone of modern traffic perception technologies.

Vision technologies constitute the foundation of TSS by providing real-time traffic scene understanding and analysis. These technologies have evolved along two distinct approaches: traditional image processing methods and modern deep learning techniques. Traditional image processing methods rely on manually designed algorithms (e.g., SIFT and SURF) to extract predefined features from images. While effective for basic tasks, these methods often struggle with complex real-world scenarios. In contrast, deep learning approaches, particularly those based on convolutional neural networks (CNNs) \cite{he2016deep} and Vision Transformer (ViT) \cite{alexey2020image}, represent a significant advancement in vision technologies. These models automatically learn to extract and analyze complex visual patterns directly from raw data, eliminating the need for hand-crafted features. The superiority of deep learning methods in TSS applications stems from their enhanced adaptability to challenging conditions (varying lighting, weather, and occlusions) and relatively robust performance in complex scenarios. These advantages have established deep learning as the predominant approach in modern TSS development.

Existing deep learning-based vision techniques in TSS generally operate at two distinct levels of traffic perception: low-level and high-level tasks. At the foundational level, low-level perception handles basic but crucial tasks such as object detection, classification, and tracking to extract fundamental information about traffic elements, including their location, category, and movement patterns. Building upon this foundation, high-level perception focuses on understanding more challenging traffic scenarios and behaviors through sophisticated applications like traffic parameter estimation, anomaly detection, and behavior understanding. These advanced tasks rely heavily on data gathered from low-level tasks, such as trajectories. Recently, the integration of foundation models, such as Large Language Models (LLMs, e.g., ChatGPT 3.5), Large Vision Models (LVMs, e.g., Segment Anything Model \cite{kirillov2023segment}) and Vision-Language Models (VLMs, e.g., CLIP \cite{radford2021learning}, GPT-4V), has opened new possibilities for achieving even more accurate and sophisticated high-level traffic perception, analysis and comprehension.

While significant scholarly attention has resulted in numerous review papers on vision-based TSS \cite{datondji2016survey,santhosh2020anomaly,boukerche2021object,espinosa2020detection,liu2020vision,zhang2022monocular,ghahremannezhad2023object,jimenez2022multi}, a comprehensive analysis bridging different perception levels and incorporating the latest technological advancements remains a critical gap. Existing surveys often exhibit limitations: (1) \textit{Fragmented Scope}: Many adopt a narrow focus, concentrating either on low-level tasks like detection and tracking \cite{boukerche2021object,ghahremannezhad2023object,jimenez2022multi} or specific high-level applications such as traffic anomaly detection  \cite{santhosh2020anomaly}, failing to provide a holistic view of the interconnected TSS pipeline. (2) \textit{Lack of Methodological Depth}: Current reviews frequently lack detailed comparative analysis of methodological approaches within task categories, limiting insights into the relative strengths and weaknesses of different techniques. (3) \textit{Omission of Emerging Technologies}: The transformative potential of foundation models (e.g., LLMs, LVMs, VLMs) in enhancing high-level traffic perception and understanding is often inadequately addressed or entirely omitted.

Our survey directly addresses these gaps and offers unique contributions by: (a) \textit{Providing a Unified Framework}: We systematically examine both low-level and high-level perception tasks within a cohesive structure, bridging the gap left by fragmented reviews. (b) \textit{Emphasizing Methodological Analysis}: We offer detailed taxonomies and comparative performance analyses for each task category, evaluating the advantages and limitations of state-of-the-art approaches, thereby providing deeper technical insights than typically found. (c) \textit{Investigating Foundation Models}: We dedicate significant focus to an in-depth investigation of foundation models, exploring their specific capabilities and potential to revolutionize TSS, a critical area often overlooked. Through this comprehensive and methodologically rigorous approach, our paper aims to deliver a more holistic and forward-looking perspective on the field. In summary, the main contributions of this paper are as follows:  

\begin{enumerate}[]
\item We conduct an in-depth investigation of foundation models in traffic perception, analyzing their distinctive capabilities (e.g., zero-shot learning, semantic understanding, and scene generation) and their transformative potential in advancing TSS applications.

\item Building upon the analysis of current TSS techniques and application limitations, we develop a systematic roadmap that identifies critical challenges and proposes specific technical innovations for future development, offering practical guidance for both researchers and practitioners.

\item We provide a comprehensive review of vision-based tasks in TSS (up to 2025), categorizing them into low-level and high-level tasks. For each category, we present a detailed methodological taxonomy, performance analysis of state-of-the-art approaches, and an evaluation of their advantages and limitations.
\end{enumerate}

\begin{figure}[!tb]
  \centering
  \includegraphics[width=\textwidth]{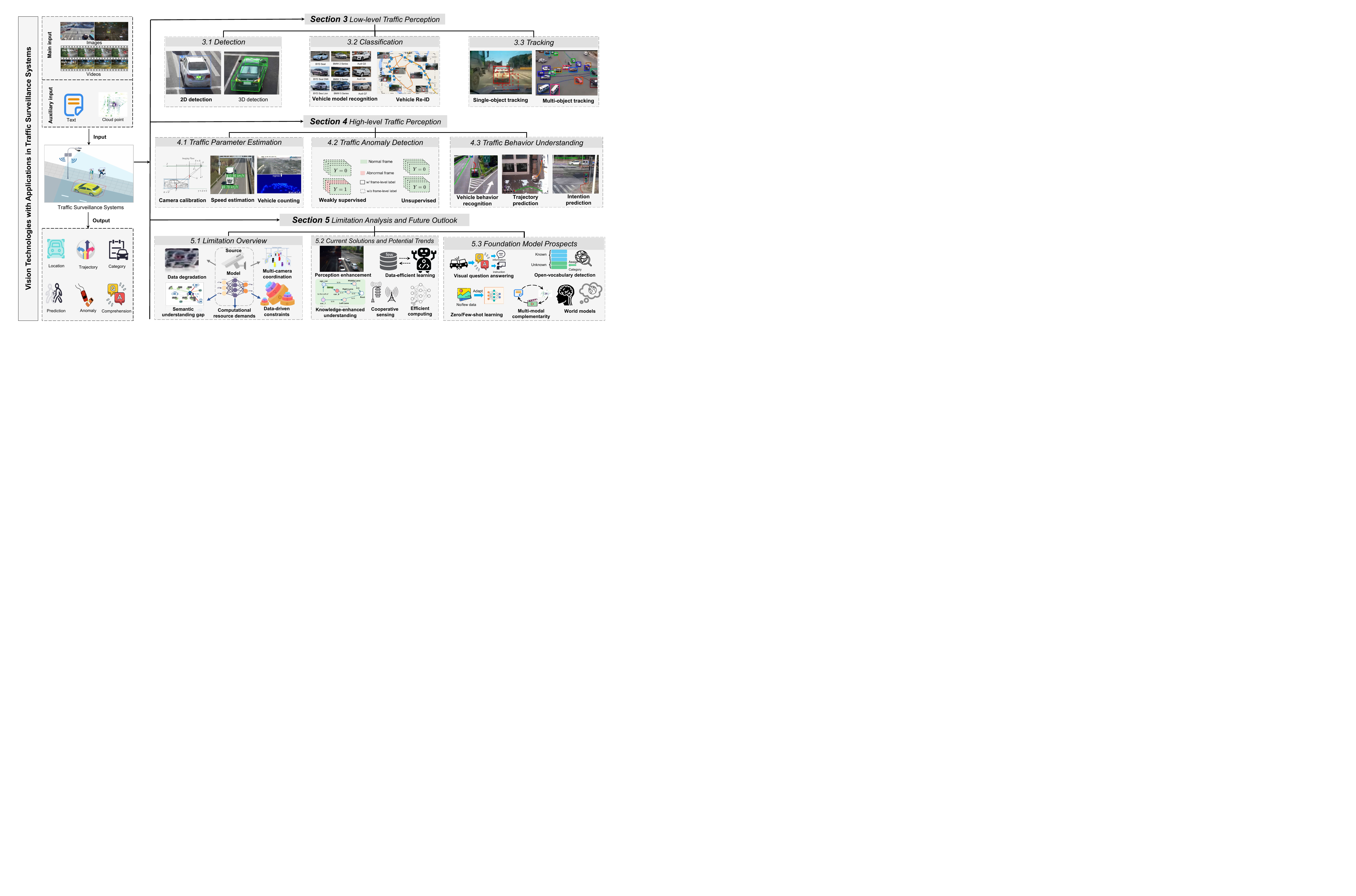}
  \caption{ Overview of vision-based TSS: core components and future prospects}
  \label{fig1}
\end{figure}

\section{Overview}

This paper is organized into three main sections that progressively explore the application of vision technologies in TSS, as illustrated in Figure \ref{fig1}. Section 3 focuses on \textit{Low-level Traffic Perception Tasks}, covering three fundamental aspects: 2D/3D detection, classification (including vehicle model recognition and vehicle Re-ID), and tracking (encompassing both single-object and multi-object tracking). Section 4 examines \textit{High-level Traffic Perception Tasks} through three advanced categories: parameter estimation (including camera calibration, speed estimation, and vehicle counting), anomaly detection (covering weakly supervised and unsupervised approaches), and behavior understanding (comprising vehicle behavior recognition, vehicle/pedestrian trajectory prediction, and intention prediction). Section 5, \textit{Limitation Analysis and Future Outlook}, first analyzes the limitations of current vision technologies in TSS scenarios, then reviews current solutions and potential trends from advanced perception technologies addressing these constraints, and concludes with future prospects centered on the distinctive capabilities of foundation models, including zero/few-shot learning, open-vocabulary detection, visual question answering, multimodal complementarity, and physical scene reasoning through world models.

\section{Low-level Traffic Perception Tasks}
In TSS, low-level traffic perception encompasses three key tasks: detection, classification, and tracking. These tasks are fundamental in obtaining essential attributes of traffic elements, such as their location, category, and trajectory. 

\subsection{Detection}
\begin{figure}[!tb]
  \centering
  \includegraphics[width=0.8\textwidth]{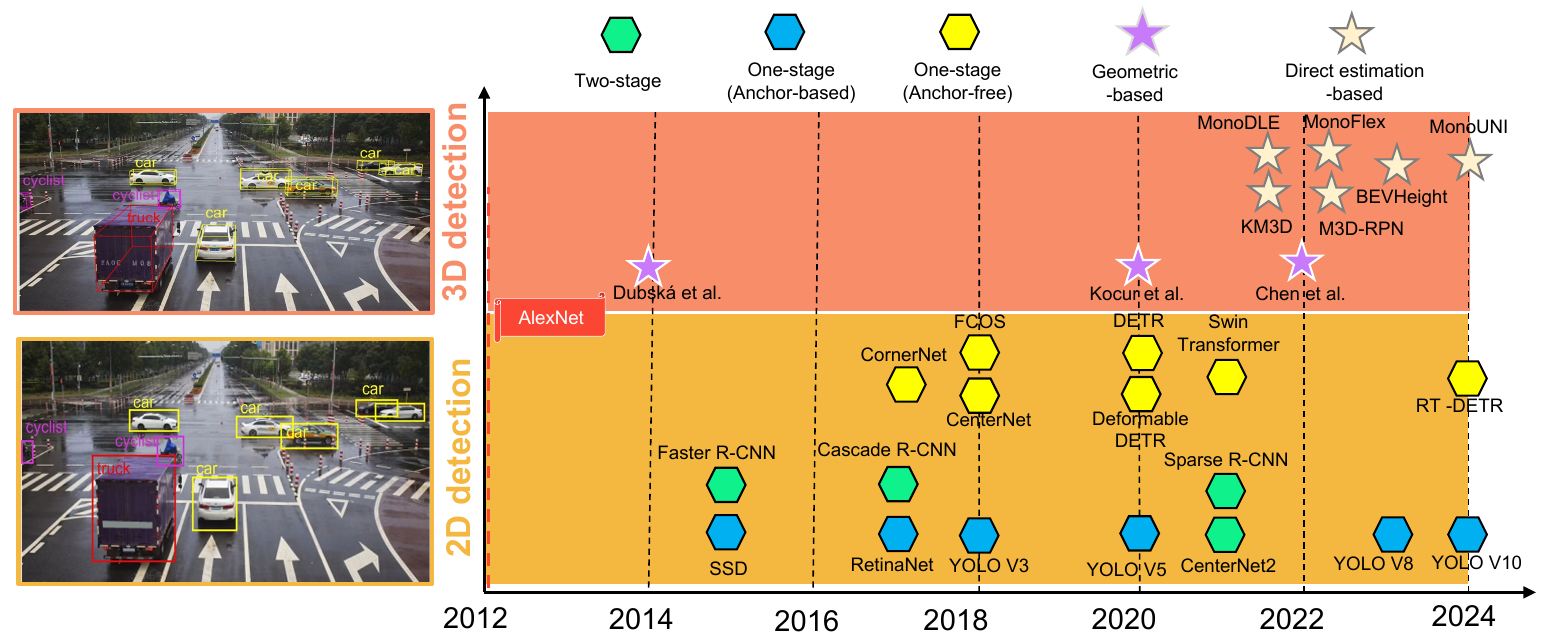}
  \caption{Evolution and categorization of mainstream methods for 2D/3D detection}
  \label{fig:2d3d}
\end{figure}

In TSS, detection involves identifying and localizing traffic elements within visual data. As shown in Figure \ref{fig:2d3d}, this process typically involves drawing either two-dimensional (2D) or three-dimensional (3D) bounding boxes around objects while assigning category labels.

\subsubsection{2D Detection}
Modern 2D detection algorithms in TSS can be categorized as \textit{two-stage} \cite{zhou2022automated,li2024toward} or \textit{one-stage} detectors \cite{zhou2023monitoring,kang2024yolo,liu2024mdfd2}, each with distinct performance characteristics and deployment considerations.

\textit{Two-stage} detectors \cite{zhou2022automated,li2024toward} (Faster R-CNN \cite{ren2016faster}, Cascade R-CNN \cite{cai2018cascade}) achieve higher accuracy in complex traffic scenes with diverse object sizes but at the cost of inference speed. Their sequential proposal-then-classification approach introduces latency (typically 2-3× slower than one-stage methods), making them less suitable for real-time applications with limited computational resources. However, they remain valuable for offline video analysis and scenarios where detection quality outweighs speed requirements, particularly for small objects like distant vehicles or pedestrians in dense urban environments \cite{zou2023object}.

\textit{One-stage} detectors prioritize efficiency through simultaneous box prediction and classification. Anchor-based methods (YOLO series \cite{redmon2016you,redmon2018yolov3,wang2024yolov10}, SSD \cite{liu2016ssd}) offer mature implementations with efficient inference in standard conditions but show performance degradation in scenes with unusual object orientations, extreme scale variations, or densely packed objects due to their reliance on predefined anchor configurations. In contrast, anchor-free approaches provide greater flexibility for diverse object shapes. CNN-based variants (FCOS \cite{tian2019fcos}, CenterNet \cite{zhou2019objects}) offer better speed-accuracy trade-offs for edge devices, while transformer-based detectors (DETR \cite{carion2020end}, RT-DETR \cite{zhao2024detrs}) demonstrate superior detection quality for complex scenes but require  more computational resources.

Current development trends reveal a clear evolution from accuracy-focused two-stage approaches toward efficiency-oriented architectures, with transformer-based methods emerging as a promising middle ground that balances detection quality and computational demands. However, persistent challenges remain in achieving robust performance across diverse traffic conditions, maintaining accuracy for small-scale objects in real-time scenarios, and developing unified architectures that can adapt to varying computational constraints \cite{zhou2023monitoring,zhou2023all,zhou2025data}.

\subsubsection{3D Detection}

3D detection in TSS generates bounding boxes reflecting objects' real-world positions. While vehicle-mounted approaches have advanced significantly, surveillance camera-based systems face distinct challenges requiring critical evaluation of methodology trade-offs. Current approaches fall into \textit{geometric-based} and \textit{direct estimation-based} categories \cite{zwemer20223d}, each with significant performance implications in real-world settings.

\textit{Geometric-based} methods leverage scene constraints to determine object depth and orientation. Approaches using vanishing points \cite{dubska2014automatic}, image transformation with 2D detection \cite{kocur2020detection}, or homography mapping \cite{chen2022monocular} can achieve reasonable accuracy in idealized conditions. However, their real-world performance varies dramatically with scene characteristics. These methods excel in structured environments with clear lane markings but degrade significantly in complex intersections or scenes with occlusion. Their sensitivity to calibration errors makes them unsuitable for dynamic surveillance environments where cameras may shift or when deployed across heterogeneous camera networks.

\begin{figure}[!tb]
  \centering
  \includegraphics[width=0.8\textwidth]{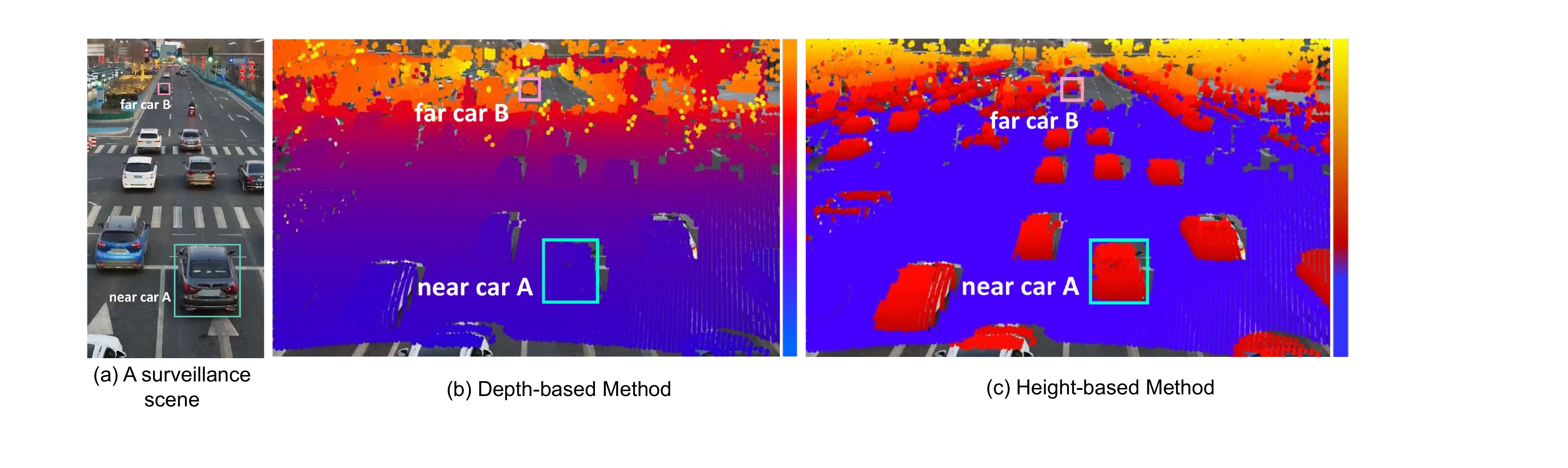}
  \caption{Depth-based methods fall short in accurately detecting vehicles that are either moving at high speeds or are situated far from the camera. In contrast, height-based methods can effectively address these limitations \cite{yang2023bevheight}}
  \label{fig4}
\end{figure}

\textit{Direct estimation-based} methods utilize deep learning to predict 3D attributes directly from images, potentially offering greater robustness and bypassing explicit calibration needs in end-to-end models, given sufficient data. This category includes adaptations of autonomous driving models like KM3D \cite{li2021monocular} for surveillance \cite{zwemer20223d}, and benchmarks such as Rope3D \cite{ye2022rope3d} evaluating methods like M3D-RPN \cite{brazil2019m3d}, MonoDLE \cite{ma2021delving}, and MonoFlex \cite{zhang2021objects}. Nevertheless, these methods heavily depend on large-scale datasets with accurate 3D annotations, which are scarce for diverse surveillance scenarios, and can be computationally intensive and less interpretable than geometric techniques. Recent advances focus on mitigating specific issues, such as height-based methods improving depth estimation \cite{yang2023bevheight} (Figure \ref{fig4})   or unified models like MonoUNI \cite{jinrang2024monouni} handling vehicle and infrastructure detection via normalized depth optimization.

\subsection{Classification}
Classification in TSS extends beyond basic categorization to fine-grained vehicle model recognition and cross-camera vehicle re-identification (Re-ID), as shown in Figure \ref{fig5}.

\subsubsection{Vehicle model recognition}

\begin{figure}[!tb]
\centering
\includegraphics[width=\textwidth]{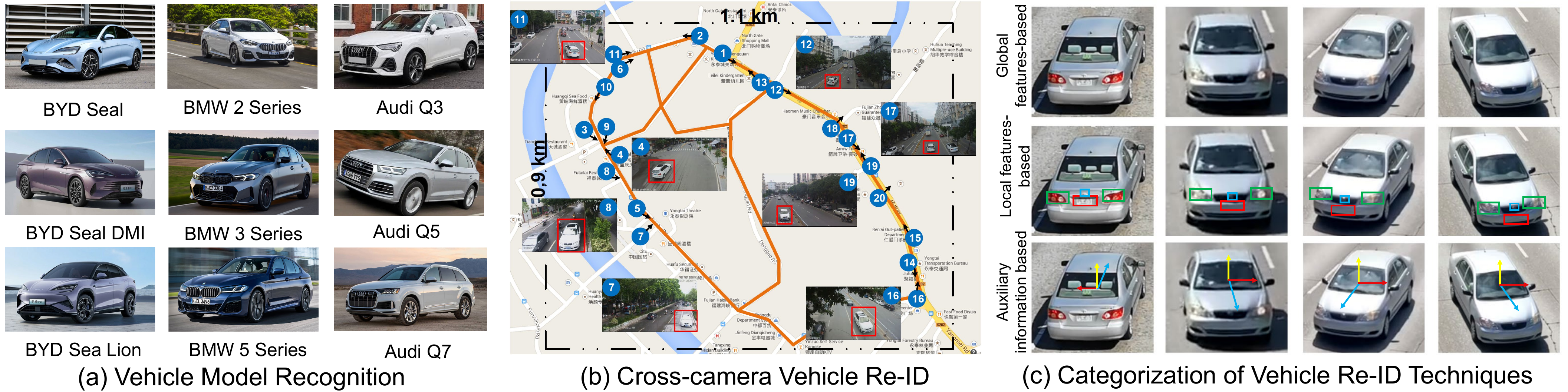}
\caption{Schematic diagram of (a) vehicle model recognition; (b) cross-camera vehicle re-identification; (c) categorization of vehicle Re-ID techniques.}
\label{fig5}
\end{figure}

Classification in TSS extends beyond basic categorization to fine-grained vehicle model recognition and cross-camera vehicle re-identification (Re-ID), as shown in Figure \ref{fig5}.

Similarly, VLR evolved from handcrafted features (SIFT, HOG, LBP) \cite{ou2014vehicle, chen2015vehicle, yu2018vehicle} to deep learning models \cite{huang2015vehicle, soon2018hyper, li2024new}. The critical trade-off here involves detection accuracy versus computational efficiency. Traditional approaches work adequately in controlled environments with frontal views but fail dramatically when logos appear at oblique angles or under partial occlusion. Deep learning models demonstrate greater robustness to these variations but at considerably higher computational cost. Hybrid approaches like MLPNL \cite{yu2020multilayer} offer a pragmatic middle ground, balancing reasonable accuracy with lower resource requirements, making them particularly suitable for deployments with limited processing capabilities.

\subsubsection{Vehicle re-identification}

Vehicle re-identification (Re-ID) matches vehicles across different camera views, with methods categorized as \textit{global feature-based}, \textit{local feature-based}, or \textit{auxiliary information-based} (Figure \ref{fig5} (c)).

\textit{Global feature-based} methods \cite{li2017deep, zhang2017improving} extract holistic vehicle representations. Their architecture simplicity enables efficient implementation and faster inference times, making them suitable for systems where computational resources are constrained. However, they struggle with fine-grained discrimination between similar vehicle models and colors—a critical limitation in real-world scenarios where many vehicles share near-identical appearances.

\textit{Local feature-based} approaches \cite{liu2018ram, huang2023coarse, lian2023multi} address this by focusing on discriminative vehicle parts. These methods achieve higher accuracy when distinctive features (license plates, roof racks, stickers) are visible. However, their performance becomes inconsistent across different camera networks with varying angles and resolutions. Combined approaches like GiT \cite{shen2023git} attempt to mitigate these limitations by integrating both perspectives but introduce additional model complexity.

\textit{Auxiliary information-based} methods incorporate viewpoint-specific metrics (VANet \cite{chu2019vehicle}), orientation conditioning \cite{khorramshahi2019dual}, attribute extraction (AttributeNet \cite{quispe2021attributenet}), or semantic information (SOFCT \cite{yu2023semantic}). While these demonstrate superior performance in challenging cross-camera scenarios with extreme viewpoint and lighting variations, their real-world applicability is constrained by the availability of additional annotations during training and potential domain gaps when deployed across different traffic surveillance networks. The performance advantage of these methods diminishes significantly when auxiliary information is limited or unavailable, raising concerns about their generalizability across diverse deployment settings.

\subsection{Tracking}
Tracking in TSS monitors objects' movement across video frames, divided into Single-Object Tracking (SOT) and Multiple-Object Tracking (MOT) approaches.

\subsubsection{Single-object tracking}
\begin{figure}[!tb]
  \centering
  \includegraphics[width=0.9\textwidth]{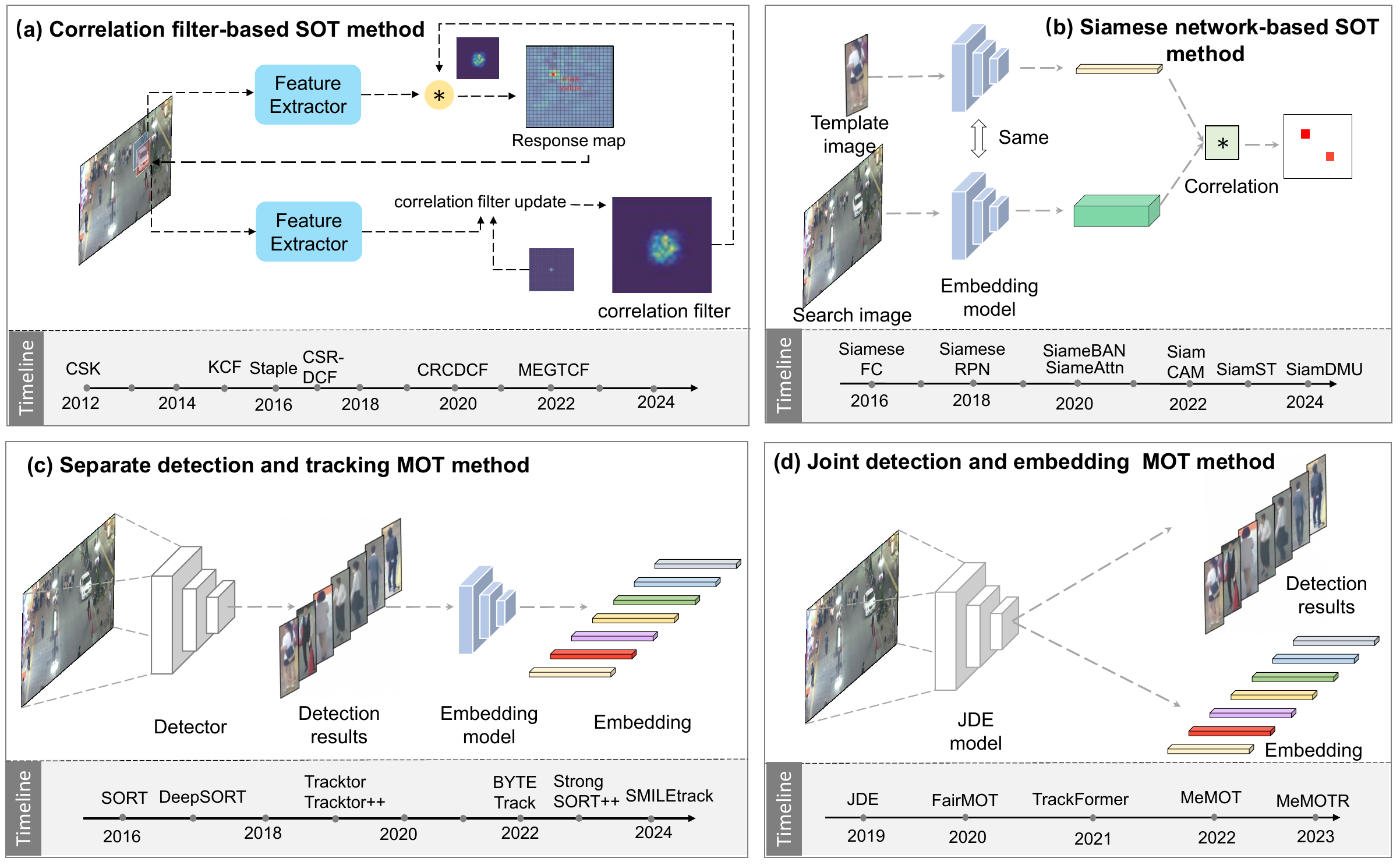}
  \caption{Pipeline and timeline of methodological development for (a) Correlation filter-based SOT methods; (b) Siamese network-based SOT methods; (c) Separate detection and tracking MOT methods; (d) Joint detection and embedding MOT methods.}
  \label{fig8}
\end{figure}

SOT methods follow a specific target throughout a video sequence, primarily using \textit{correlation filter-based} or \textit{siamese network-based} approaches (Figure \ref{fig8} (a-b)).

\textit{Correlation filter-based} methods (\cite{bolme2010visual, henriques2014high, bertinetto2016staple}) update appearance filters iteratively, offering computational efficiency suitable for resource-constrained surveillance hardware. However, they demonstrate significant performance degradation in common traffic scenarios with occlusions and illumination variations. Their adaptive nature leads to error accumulation during extended tracking, making them less reliable for prolonged surveillance without periodic reinitialization.

\textit{Siamese network-based} approaches \cite{bertinetto2016fully} employ deep learning for similarity measurement, with advancements in region proposals \cite{li2018high}, attention mechanisms \cite{yu2020deformable}, and spatiotemporal modeling \cite{zhang2023siamst}. They show superior robustness to appearance changes but require substantial computational resources, limiting real-time performance on edge devices. Their accuracy drops  when tracking small, distant objects or in adverse weather conditions absent from training data.

\subsubsection{Multi-object tracking}

Multiple Object Tracking (MOT) simultaneously tracks multiple targets in video sequences, essential for vehicle and pedestrian tracking in TSS. MOT methods are categorized into Separate Detection and Tracking (SDT) and Joint Detection and Embedding (JDE) paradigms, as shown in Figure \ref{fig8}(c-d).

The SDT paradigm operates through object detection, feature extraction, and cross-frame tracking. SORT \cite{bewley2016simple} combines Kalman filtering with Hungarian algorithm, while DeepSORT \cite{wojke2017simple} adds deep feature representations. Recent advances include BYTETrack's \cite{zhang2022bytetrack} two-stage association, StrongSORT++ \cite{du2023strongsort}'s multi-aspect improvements, and SMILEtrack \cite{wang2024smiletrack}'s self-attention mechanisms. However, this multi-stage process often incurs higher computational costs in terms of both processing time (latency/lower FPS) and memory usage (particularly GPU memory), as it involves running distinct models/algorithms sequentially. Furthermore, this approach is prone to error propagation from the detector.

The JDE paradigm integrates detection and tracking in a unified framework. Following Wang et al.'s \cite{wang2020towards} pioneering JDE model, FairMOT \cite{zhang2021fairmot} enhanced feature extraction through Deep Layer Aggregation. Recent Transformer-based approaches like TrackFormer \cite{meinhardt2022trackformer} and MeMOTR \cite{gao2023memotr} utilize self-attention for improved inter-target relationship modeling. While computationally efficient due to shared backbones and end-to-end training, this paradigm inherently offers less flexibility. The joint optimization can lead to a compromise where the learned features may not be optimal for either detection or re-identification (embedding) tasks individually, potentially capping the performance ceiling for each.

\subsection{Performance Evaluation}
The evaluation of low-level perception tasks in TSS relies on comprehensive datasets and specialized metrics for each task. This section first details the datasets and evaluation metrics used for detection, classification, and tracking tasks, and then presents the results of some representative approaches.

\begin{table*}[htbp]  
  \centering  
\makeatletter\setlength{\@fptop}{0pt}\makeatother  
\hspace*{-2cm}  
\begin{minipage}{\dimexpr\textwidth+3cm\relax}  
  \centering  
\renewcommand\arraystretch{0.1} % Assuming this very small stretch is intentional
  \caption{Overview of common datasets for 2D/3D detection, fine-grained vehicle classification (FGVC), vehicle logo recognition (VLR), vehicle Re-ID and single/multiple object tracking, including sensor details and TSS relevance. N/A denotes “Not applicable”. Size units: \textcolor{blue}{I} (Images), \textcolor{orange}{V} (Videos), \textcolor{green}{O} (Objects).} 
    \begin{tabular}{>{\centering\arraybackslash}m{1.5cm}|>{\centering\arraybackslash}m{2.7cm}|>{\centering\arraybackslash}m{0.7cm}|>{\centering\arraybackslash}m{1.5cm}|>{\centering\arraybackslash}m{1.5cm}|>{\centering\arraybackslash}m{2.8cm}|>{\centering\arraybackslash}m{6cm}|>{\arraybackslash}m{1cm}}  
    \toprule  
    \toprule  
    \multicolumn{1}{c|}{\multirow{4}[2]{*}{\textbf{Task}}} & \multirow{4}[2]{*}{\textbf{Dataset }} & \multicolumn{1}{c|}{\multirow{4}[2]{*}{\textbf{Year}}} & \textbf{Size} & \multirow{4}[2]{*}{\textbf{Class Num.}} & \multirow{4}[2]{*}{\textbf{Sensor Platform}} & \multirow{4}[2]{*}{\textbf{Key Characteristics \& TSS Focus}} & \multirow{4}[2]{*}{\textbf{Link}} \\
          & \multicolumn{1}{c|}{} &       & \textbf{(\textcolor{blue}{I}; \textcolor{orange}{V}; \textcolor{green}{O})} & \multicolumn{1}{c|}{} & \multicolumn{1}{c|}{} & \multicolumn{1}{c|}{} & \multicolumn{1}{c}{} \\ % Adjusted size legend to fit
          & \multicolumn{1}{c|}{} &       & & \multicolumn{1}{c|}{} & \multicolumn{1}{c|}{} & \multicolumn{1}{c|}{} & \multicolumn{1}{c}{} \\
          & \multicolumn{1}{c|}{} &       & & \multicolumn{1}{c|}{} & \multicolumn{1}{c|}{} & \multicolumn{1}{c|}{} & \multicolumn{1}{c}{} \\
    \midrule  
    \midrule  
    \multicolumn{1}{c|}{\multirow{5}[20]{*}{2D Detection}} & UA-DETRAC \cite{wen2020ua} & 2015  & 140,000+ \textcolor{blue}{I} & \multicolumn{1}{c|}{4} & Static (Surveillance-like, Roadside, China) & Varied illumination/occlusion; Robust vehicle detection/MOT  & {\small \href{https://universe.roboflow.com/rjacaac1/ua-detrac-dataset-10k}{Link}} \\
\cmidrule{2-8}          & Freeway-Vehicle \cite{song2019vision} & 2019  & 11,129 \textcolor{blue}{I} & \multicolumn{1}{c|}{3} & Static (Freeway surveillance, China) & High-speed and low-density freeway scenarios;  Vehicle detection & {\small \href{https://drive.google.com/open?id=1li858elZvUgss8rC_yDsb5bDfiRyhdrX}{Link}} \\
\cmidrule{2-8}          & \multirow{2}[2]{*}{MIO-TCD \cite{luo2018mio}} & \multirow{2}[2]{*}{2018} & \multirow{2}[2]{*}{786,702 \textcolor{blue}{I}} & \multicolumn{1}{c|}{\multirow{2}[2]{*}{11}} & Static (Traffic surveillance, Canada/USA) & Diverse urban object classes; Multi-class vehicle detection & {\small \href{https://tcd.miovision.com/challenge/dataset.html}{Link}} \\
          & \multicolumn{1}{c|}{} &       & \multicolumn{1}{c|}{} & \multicolumn{1}{c|}{} &  &  & \multicolumn{1}{c}{} \\
\cmidrule{2-8}          & SEU\_PML \cite{zhou2023monitoring} & 2023  & 270,000 \textcolor{green}{O} & \multicolumn{1}{c|}{13} & Static (Traffic surveillance, China) & High-density urban scenarios with high-quality annotation; Traffic participant detection &  {\small \href{https://github.com/vvgoder/SEU\_PML\_Dataset}{Link}}\\
    \midrule  
    \multicolumn{1}{c|}{\multirow{5}[30]{*}{3D Detection}} & BAAI-VANJEE \cite{yongqiang2021baai} & 2021  & 7,500 \textcolor{blue}{I} & \multicolumn{1}{c|}{12} & Static (Infrastructure LiDAR/Cam, China) & Challenging roadside infrastructure views in Chinese traffic; Roadside 3D detection, V2I applications & {\small \href{https://data.baai.ac.cn/data-set}{Link}} \\
\cmidrule{2-8}          & IPS300+ \cite{wang2022ips300+} & 2022  & 14,198 \textcolor{blue}{I} & \multicolumn{1}{c|}{7} & Static (Intersection Perception System, China) & Crowded urban intersection covering area of 3000m²; V2I, safety analysis & {\small \href{http://openmpd.com/column/IPS300}{Link}} \\
\cmidrule{2-8}          & A9-dataset \cite{cress2022a9} & 2022  & 1,098 \textcolor{blue}{I} & \multicolumn{1}{c|}{9} & Static (Roadside Infrastructure LiDAR \& Camera, Germany) & High-speed dynamic 3D scenes (motorway); Autonomous driving and V2I applications & {\small \href{https://a9-dataset.com}{Link}} \\
\cmidrule{2-8}          & Rope3D \cite{ye2022rope3d} & 2022  & 50k+ \textcolor{blue}{I} & \multicolumn{1}{c|}{13} & Static (Roadside Cam/LiDAR, China) & First high-diversity challenging roadside 3D dataset;  Long-range 3D  detection & {\small \href{https://thudair.baai.ac.cn/rope}{Link}} \\
\cmidrule{2-8}          & DAIR-V2X \cite{yu2022dair} & 2022  & 71,254 \textcolor{blue}{I} & \multicolumn{1}{c|}{10} & Static (Roadside); Moving (Vehicle) LiDAR/Cam (China) & First real-world vehicle-infrastructure collaborative dataset; Cooperative 3D perception (V2X) & {\small \href{https://github.com/AIR-THU/DAIR-V2X}{Link}} \\
    \midrule  
    \multicolumn{1}{c|}{\multirow{2}[5]{*}{FGVC}} & Stanford Cars \cite{krause20133d} & 2013  & 16,185 \textcolor{blue}{I} & \multicolumn{1}{c|}{196} & Static Viewpoints (General) & Many car classes, often photographed from the rear; Vehicle model recognition &  {\small \href{https://huggingface.co/datasets/tanganke/stanford_cars}{Link}} \\
\cmidrule{2-8}          & CompCars \cite{yang2015large} & 2015  & 30,955 \textcolor{blue}{I} & \multicolumn{1}{c|}{431} & Static Viewpoints (Internet \& Surveillance Cams) & Rich vehicle attribute data (e.g., max speed, displacement, doors, seats, type); Vehicle attribute recognition & {\small \href{https://mmlab.ie.cuhk.edu.hk/datasets/comp\_cars/}{Link}} \\
    \midrule  
    \multicolumn{1}{c|}{\multirow{3}[8]{*}{VLR}} & HFUT-VL \cite{yu2018vehicle} & 2018  & 32,000 \textcolor{blue}{I} & \multicolumn{1}{c|}{80} & Static (Traffic surveillance) & Real-world surveillance imagery; Vehicle logo recognition, manufacturer identification & {\small \href{https://github.com/HFUT-VL/HFUT-VL-dataset}{Link}} \\
\cmidrule{2-8}          & XMU \cite{huang2015vehicle} & 2015  & 11,500 \textcolor{blue}{I} & \multicolumn{1}{c|}{10} & Static (Traffic surveillance) & Early benchmark with major manufacturers; Vehicle manufacturer recognition & {\small \href{https://smartdsp.xmu.edu.cn/}{Link}} \\
\cmidrule{2-8}          & VLD-45 \cite{yang2021vld} & 2022  & 45,000 \textcolor{blue}{I} & \multicolumn{1}{c|}{45} & Static Viewpoints (Internet) & Variable logo sizes and perspectives; Vehicle logo recognition and detection & {\small \href{https://github.com/YangShuoys/VLD-45-B-DATASET-Detection}{Link}} \\
    \midrule  
    \multicolumn{1}{c|}{\multirow{4}[14]{*}{Vehicle Re-ID}} & VehicleID \cite{liu2016deep} & 2016  & 221,763 \textcolor{blue}{I} & 250   & Static (Traffic surveillance) & Real-world surveillance data captured  from  a small Chinese city; Large-scale Re-ID & {\small \href{https://pkuml.org/resources/pku-vehicleid.html}{Link}} \\
\cmidrule{2-8}          & VeRI-776 \cite{liu2016deepb} & 2016  & 49,357 \textcolor{blue}{I} & N/A   & Static (Networked surveillance cams) &  Viewpoint/illumination changes with spatio-temporal annotations; Cross-camera Re-ID & {\small \href{https://github.com/JDAI-CV/VeRidataset}{Link}} \\
\cmidrule{2-8}          & CityFlow \cite{tang2019cityflow} & 2019  & 229,680 \textcolor{blue}{I} & N/A   & Static (Networked surveillance cams, US) & City-scale multi-camera Re-ID;  Large-scale vehicle Re-ID & {\small \href{https://www.aicitychallenge.org/2020-data-access-instructions/}{Link}} \\
\cmidrule{2-8}          & VERI-Wild 2.0 \cite{bai2021disentangled} & 2021  & 825,042 \textcolor{blue}{I} & N/A   & Static (Unconstrained surveillance cams) & Unconstrained environments with extreme appearance variations; Robust re-identification & {\small \href{https://github.com/PKU-IMRE/VERI-Wild}{Link}} \\
    \midrule  
    \multicolumn{1}{c|}{\multirow{2}[4]{*}{SOT}} & UAV123 \cite{benchmark2016benchmark} & 2016  & 123 \textcolor{orange}{V} & N/A   & Moving (Drone/UAV cameras) & Aerial perspectives with long sequences; Aerial SOT & {\small \href{https://datasetninja.com/uav123}{Link}} \\
\cmidrule{2-8}          & VisDrone-SOT \cite{zhu2018vision} & 2019  & 157 \textcolor{orange}{V} & N/A   & Moving (Drone/UAV cameras) & Complex urban environments with small and occluded objects; City-wide SOT & {\small \href{https://github.com/VisDrone/VisDrone-Dataset}{Link}} \\
    \midrule  
    \multicolumn{1}{c|}{\multirow{3}[8]{*}{MOT}} & UA-DETRAC \cite{wen2020ua} & 2020  & 100 \textcolor{orange}{V} & N/A   & Static (Surveillance-like, Roadside, China) & Varied illumination/occlusion; Robust vehicle detection/MOT benchmark & {\small \href{https://universe.roboflow.com/rjacaac1/ua-detrac-dataset-10k}{Link}} \\
\cmidrule{2-8}          & VisDrone-MOT \cite{zhu2018vision} & 2019  & 79 \textcolor{orange}{V} & N/A   & Moving (Drone/UAV cameras, China) & Dense object distribution with diverse scenes; UAV-based and city-wide MOT & {\small \href{https://github.com/VisDrone/VisDrone-Dataset}{Link}} \\
\cmidrule{2-8}          & MOT20 \cite{dendorfer2021motchallenge} & 2020  & 8 \textcolor{orange}{V} & N/A   & Static \& Moving Cams (Challenging MOT benchmark) & Extremely crowded scenes; Robust tracking in dense crowds & {\small \href{https://motchallenge.net/data/MOT20/}{Link}} \\
    \bottomrule  
    \end{tabular}%  
    % \vspace{1em} 
    % \begin{justify} 
    % \small 
    % \textbf{Note:} FGVC is short for Fine-grained Vehicle Classification, VLR is short for Vehicle Logo Recognition. Size units: \textcolor{blue}{I} (Images), \textcolor{orange}{V} (Videos), \textcolor{green}{O} (Objects). Table entries provide concise summaries.
    % \end{justify}  
  \label{tab1}%  
  \end{minipage}
\end{table*}%

\subsubsection{Datasets for low-level perception}

In terms of \textbf{\textit{detection}} tasks in TSS, representative datasets include UA-DETRAC \cite{wen2020ua}, Freeway-Vehicle \cite{song2019vision}, MIO-TCD \cite{luo2018mio}, and SEU\_PML \cite{zhou2023monitoring} for 2D detection, as well as BAAI-VANJEE \cite{yongqiang2021baai}, IPS300+ \cite{wang2022ips300+}, A9-dataset \cite{cress2022a9}, Rope3D \cite{ye2022rope3d}, and DAIR-V2X \cite{yu2022dair} for 3D detection.

In terms of \textbf{\textit{classification}} tasks in TSS, representative datasets include Stanford Cars \cite{krause20133d} and CompCars \cite{yang2015large} for fine-grained vehicle classification, HFUT-VL \cite{yu2018vehicle}, XMU \cite{huang2015vehicle}, and VLD-45 \cite{yang2021vld} for vehicle logo recognition, as well as VehicleID \cite{liu2016deep}, VeRI-776 \cite{liu2016deepb}, CityFlow \cite{tang2019cityflow}, and VERI-Wild 2.0 \cite{bai2021disentangled} for vehicle Re-ID.

In terms of \textbf{\textit{tracking}} tasks in TSS, representative datasets include UAV123 \cite{benchmark2016benchmark} and VisDrone-SOT \cite{zhu2018vision} for single object tracking (SOT), as well as UA-DETRAC \cite{wen2020ua}, MOT \cite{dendorfer2021motchallenge}, and VisDrone-MOT2019 \cite{zhu2018vision} for multiple object tracking (MOT). More detailed statistics is shown in Table \ref{tab1}.

\subsubsection{Metrics and performance evaluation}
Evaluation metrics for 2D and 3D detection share similar principles while differing in implementation details. For 2D object detection, commonly used metrics include IOU (Intersection Over Union), Precision, Recall, F1 Score, Average Precision (AP), and Mean Average Precision (mAP) \cite{zou2023object}.

For 3D detection, similar metrics are adapted with 3D mAP and BEV (Bird's Eye View) mAP being calculated using 3D IOU or BEV IOU respectively \cite{song2024robustness}. However, the evaluation emphasis differs significantly based on sensor modality: \textit{LiDAR-based 3D methods}, benefiting from direct 3D point cloud measurements, are evaluated with strict focus on geometric precision, typically using higher IoU thresholds (e.g., 0.7). In contrast, \textit{image-based 3D methods} (e.g., monocular or stereo) must infer depth and 3D structure, leading to evaluations that assess both 3D detection accuracy and depth estimation quality, often with more lenient IoU thresholds (e.g., 0.5) to account for inherent depth inference uncertainty. Notably, different domains employ specialized evaluation frameworks. For instance, the KITTI dataset \cite{geiger2013vision}, which provides benchmarks for both \textit{image-based} and \textit{LiDAR-based} 3D detection, uses the 11-point Interpolated Average Precision. The nuScenes dataset \cite{caesar2020nuscenes}, utilizing a multi-modal sensor suite including LiDAR and cameras, implements a comprehensive framework including mAP and various True Positive metrics (e.g., ATE, ASE, AOE, AVE, AAE) that capture different facets of 3D detection accuracy. In Traffic Surveillance Systems, specialized metrics have emerged, as exemplified by the Rope3D dataset’s evaluation system for image-based long-range 3D detection, which includes metrics like ACS, AOS, AAS, AGD, and AGS \cite{ye2022rope3d}.

For the task of fine-grained vehicle classification and vehicle logo recognition, common evaluation metrics primarily includes Accuracy (Acc) and Confusion Matrices (CM). For Re-ID tasks, main metrics include: RR (Rank Ratio), mAP (Mean Average Precision), and CMC (Cumulative Matching Characteristic) \cite{amiri2024comprehensive}.

SOT and MOT tracking tasks utilize different evaluations metrics suited to their specific characteristics. SOT evaluation primarily relies on four key metrics: Success Rate (measuring overlap between tracking results and ground truth), Success Plot (visualizing Success Rate across different thresholds), Average Overlap Rate (AOR, calculating mean overlap), and Expected Average Overlap (EAO, measuring overall tracking accuracy) \cite{li2018high}. For MOT, the main metrics include MOTA (evaluating overall accuracy considering missed detections, false positives, and ID switches), MOTP (assessing positional accuracy), IDF1 (measuring ID matching performance), IDs (counting identity switches), and FPS (indicating real-time processing capability) \cite{luo2021multiple}.

Table \ref{tab2} presents performance benchmarks for representative methods across key traffic perception tasks, revealing several important technological evolution patterns and research directions. For 2D detection, the results demonstrate a clear progression from two-stage methods toward more efficient one-stage approaches, with recent transformer-based methods achieving superior performance while maintaining real-time capabilities. In 3D detection, direct estimation methods significantly outperform geometric approaches, indicating that data-driven learning has largely superseded hand-crafted geometric constraints.

The classification tasks (FGVC and VLR) show consistent improvements with deep learning over handcrafted features, reaching accuracy levels above 94\% and 99\% respectively, suggesting these tasks are approaching practical deployment readiness. For vehicle Re-ID, local feature methods generally outperform global approaches, while auxiliary information integration shows promise for handling challenging scenarios. In tracking, the evolution from filter-based methods to Siamese networks in SOT, and from separate detection-tracking to joint frameworks in MOT, reflects the field's movement toward end-to-end learning paradigms.

These performance trends suggest future research should focus on: (1) unified transformer architectures that balance accuracy and efficiency across detection scales, (2) self-supervised learning approaches to reduce dependency on large annotated datasets, (3) multi-modal fusion techniques for robust performance in adverse conditions, and (4) joint optimization frameworks that integrate multiple perception tasks for computational efficiency in real-world deployments.

\section{High-level Traffic Perception Tasks}

High-level traffic perception in TSS builds upon low-level perception tasks to achieve analysis and understanding of traffic scenes. The scope of high-level perception encompasses critical tasks including traffic parameter extraction, traffic anomaly detection, and vehicle/pedestrian behavior understanding. 

\subsection{Traffic Parameter Estimation}
Traffic parameter estimation quantifies key traffic characteristics including flow rate, density, vehicle speed, and occupancy. Camera calibration remains a fundamental challenge affecting all measurement accuracy. This section examines calibration approaches and two critical estimation tasks: speed estimation and vehicle counting.

\subsubsection{Camera Calibration}
Camera calibration \cite{zhang2023camera} determines intrinsic and extrinsic parameters that enable conversion between image and world coordinates—essential for accurate measurements.

While active calibration methods \cite{basu1997active,chen2024novel} exist, TSS typically employs two more practical approaches: \textit{vanishing point-based} and \textit{vehicle keypoint-based} methods.

\textit{Vanishing point-based} methods (Figure \ref{fig:highlevel_class}(a)) utilize convergence points of parallel lines. Approaches include motion blob tracking \cite{thi2008self}, multi-source fusion \cite{zheng2012model}, vehicle trajectory analysis \cite{dubska2014automatic}, and three-VP methods \cite{orghidan2012camera,zhang2012practical}. Recent advances include CNN-based approaches \cite{kocur2021traffic}, automatic highway calibration \cite{zhang2023automatic}, and online auto-calibration \cite{guo2024online}. These methods struggle in scenes lacking clear parallel structures and show significant error propagation in calculations. Their performance degrades in urban environments with irregular road geometries and requires periodic recalibration due to sensitivity to camera movement or environmental changes.

\hspace*{-2cm}  % 向左移动2厘米
\begin{minipage}{\dimexpr\textwidth+2cm\relax}  % 创建一个比文本宽度宽4厘米的小页环境
\renewcommand{\arraystretch}{0.4} 
\footnotesize
\begin{longtable}{c|c|p{45em}|c|p{16em}}
\caption{Performance of current representative methods for 2D/3D detection, fine-grained vehicle classification, vehicle logo recognition, vehicle Re-ID and single/multiple object tracking.\label{tab2}}\\
\toprule  
\toprule  
\multicolumn{1}{p{2.5em}|}{\textbf{Task}} & \multicolumn{1}{p{3em}|}{\textbf{Category}} & \textbf{Method: \textcolor{gray}{Brief description}} & \multicolumn{1}{p{3.715em}|}{\textbf{Year}} & \textbf{Benchmark: Metrics (\textcolor{cyan}{Source})}\\
\midrule  
\midrule  
\endfirsthead  

\multicolumn{5}{c}%  
{{\bfseries \tablename\ \thetable{} -- continued from previous page}} \\
\toprule  
\toprule  
\multicolumn{1}{p{2.5em}|}{\textbf{Task}} & \multicolumn{1}{p{3em}|}{\textbf{Category}} & \textbf{Method} & \multicolumn{1}{p{3.715em}|}{\textbf{Year}} & \multicolumn{1}{p{10.715em}|}{\textbf{Benchmark: Metrics (Source)}} \\
\midrule  
\midrule  
\endhead  

\midrule  
\multicolumn{5}{r}{{Continued on next page}} \\
\endfoot  

\bottomrule  
\bottomrule  
\endlastfoot  
% 表格内容开始  
\multicolumn{1}{c|}{\multirow{13}[20]{*}{\shortstack{2D \\Detection}}} & \multicolumn{1}{c|}{\multirow{1}[2]{*}{Two-stage}} & \makecell[l]{Faster R-CNN~\cite{ren2016faster}: \textcolor{gray}{Region Proposal Network with two-stage detection pipeline}} & 2015  & \shortstack[l]{SEU\_PML: mAP =62.53\%} ({\small \href{https://ieeexplore.ieee.org/abstract/document/10226455}{Link}})    \\
\cmidrule{3-5}          &       & Cascade R-CNN \cite{cai2018cascade}: \textcolor{gray}{Series of detectors trained with increasing IoU thresholds for progressive refinement} & 2017  & SEU\_PML: mAP =65.66\% ({\small \href{https://ieeexplore.ieee.org/abstract/document/10226455}{Link}}) \\
\cmidrule{2-5}          & \multicolumn{1}{c|}{\multirow{1}[12]{*}{\shortstack{One-stage\\(Anchor-based)}}} & YOLO V3 \cite{redmon2018yolov3}: \textcolor{gray}{Single-stage detector with multi-scale feature maps and anchor boxes} & 2018  & \shortstack[l]{ SEU\_PML: mAP =61.54\%} ({\small \href{https://ieeexplore.ieee.org/abstract/document/10226455}{Link}})\\
\cmidrule{3-5}          &       & YOLO V5\footnote{https://docs.ultralytics.com/yolov5}: \textcolor{gray}{Enhanced YOLO with CSPNet backbone and PANet feature aggregation} & 2020  & \shortstack[l]{SEU\_PML: mAP =66.86\%} ({\small \href{https://github.com/ultralytics/ultralytics}{Link}})\\
\cmidrule{3-5}          &       & YOLO V8\footnote{https://github.com/ultralytics/ultralytics}: \textcolor{gray}{End-to-end model with anchor-free detection head and C2f block architecture} & 2023  & VisDrone-DET: mAP = 39.0\%  ({\small \href{https://github.com/ultralytics/ultralytics}{Link}})\\
\cmidrule{3-5}          &       & YOLO V11\footnote{https://github.com/ultralytics/ultralytics}: \textcolor{gray}{Latest YOLO iteration with a re-architected backbone and neck for enhanced feature acuity} & 2024  & UA-DETRAC: mAP =55.9\% ({\small \href{https://github.com/ultralytics/ultralytics}{Link}}) \\
\cmidrule{2-5}          & \multicolumn{1}{c|}{\multirow{5}[9]{*}{\shortstack{One-stage\\(Anchor-free)}}} &  CenterNet \cite{zhou2019objects}: \textcolor{gray}{Keypoint-based detector that predicts object centers and regresses size} & 2019  & VisDrone-DET: mAP = 38.6\% ({\small \href{https://ieeexplore.ieee.org/abstract/document/10928877}{Link}}) \\
\cmidrule{3-5}          &       & DETR \cite{carion2020end}: \textcolor{gray}{Transformer-based detector with bipartite matching and parallel decoding} & 2020  & BIT-Vehicle: mAP = 90.2\% ({\small \href{https://ieeexplore.ieee.org/document/10581739}{Link}})  \\
\cmidrule{3-5}          &       & Deformable DETR \cite{zhu2020deformable}: \textcolor{gray}{DETR with deformable attention for multi-scale feature processing} & 2020  & BIT-Vehicle: mAP = 90.8\% ({\small \href{https://ieeexplore.ieee.org/document/10581739}{Link}}) \\
\cmidrule{3-5}          &       & Swin Transformer \cite{liu2021swin}: \textcolor{gray}{Hierarchical vision transformer with shifted window attention mechanism} & 2021  &  VisDrone-DET: mAP = 63.9\% ({\small \href{https://ieeexplore.ieee.org/document/9607674}{Link}}) \\
\cmidrule{3-5}          &       & RT-DETR \cite{zhao2024detrs}: \textcolor{gray}{Real-time transformer detector with hybrid encoder and IoU-aware head} & 2024  & VisDrone-DET: mAP = 72.7\% ({\small \href{https://ieeexplore.ieee.org/abstract/document/10423771/}{Link}}) \\
\midrule  
\multicolumn{1}{c|}{\multirow{6}[16]{*}{\shortstack{3D\\ Detection}}} & \multicolumn{1}{c|}{\multirow{1}[2]{*}{Geometric}} & Dubská et al. \cite{dubska2014automatic}: \textcolor{gray}{Vehicle measurement using vanishing point analysis and 3D bounding box reconstruction} & 2014  & Proprietary: Mean Error<=2\%  ({\small \href{https://bmva-archive.org.uk/bmvc/2014/papers/paper013/index.html}{Link}})  \\
\cmidrule{3-5}          &       & Chen et al. \cite{chen2022monocular}: \textcolor{gray}{Geometric method for calibration-free 3D vehicle box estimation without 3D annotations} & 2022  & Ko-PER: AP= 70.53\% ({\small \href{https://ieeexplore.ieee.org/document/9747512}{Link}})  \\
\cmidrule{2-5}          & \multicolumn{1}{c|}{\multirow{5}[12]{*}{\shortstack{Direct \\ estimation}}} &  M3D-RPN \cite{brazil2019m3d}: \textcolor{gray}{Monocular 3D region proposal network with depth-aware convolutions} & 2022  & Rope3D dataset: AP3D =67.17\% ({\small \href{https://ieeexplore.ieee.org/document/9879696}{Link}})  \\
\cmidrule{3-5}          &       & MonoDLE \cite{ma2021delving}: \textcolor{gray}{Monocular 3D detection tackling localization errors with refined training and a  3D IoU-based loss} & 2022  & Rope3D dataset: AP3D =77.50\%  ({\small \href{https://ieeexplore.ieee.org/document/9879696}{Link}})\\
\cmidrule{3-5}          &       & MonoFlex \cite{zhang2021objects}: \textcolor{gray}{Flexible framework for 3D object detection with uncertainty modeling} & 2022  & Rope3D dataset: AP3D =59.78\%  ({\small \href{https://ieeexplore.ieee.org/document/9879696}{Link}})\\
\cmidrule{3-5}          &       & {BEVHeight \cite{yang2023bevheight}}: \textcolor{gray}{Bird’s-eye-view detector that regresses object height to guide 3D detection} & {2023} & Rope3D dataset: AP3D = 74.60\%({\small \href{https://arxiv.org/abs/2303.08498}{Link}})\\
          
\cmidrule{3-5}          &       & MonoUNI \cite{jinrang2024monouni}: \textcolor{gray}{ SOTA unified 3D (vehicles/infra) leveraging normalized depth \& 3D cube depth} & 2024  & Rope3D dataset: AP3D = 92.45\% ({\small \href{https://github.com/Traffic-X/MonoUNI}{Link}}) \\
\midrule  
\multicolumn{1}{c|}{\multirow{6}[12]{*}{FGVC}} & \multicolumn{1}{c|}{\multirow{2}[4]{*}{\shortstack{\shortstack{Handcrafted\\ features}}}} & Krause et al. \cite{krause20133d}: \textcolor{gray}{3D geometric reasoning with localized part detection for fine-grained categorization} & 2013  & Stanford Cars: Accuracy= 67.6\%  ({\small \href{https://ieeexplore.ieee.org/document/6755945}{Link}})  \\
\cmidrule{3-5}          &       & Hsieh et al. \cite{hsieh2014symmetrical}: \textcolor{gray}{Symmetry-based vehicle detection with HOG feature extraction} & 2014  & Proprietary: Precision= 98.1\%  ({\small \href{https://ieeexplore.ieee.org/document/6720121}{Link}}) \\
\cmidrule{2-5}          & \multicolumn{1}{c|}{\multirow{4}[8]{*}{\shortstack{Deep\\ learning}}} & Sochor et al. \cite{sochor2018boxcars}: \textcolor{gray}{Fine-grained vehicle recognition by “unpacking” images using 3D bounding boxes and CNN } & 2019  & Boxcars116k: Accuracy= 84.13\% ({\small \href{https://arxiv.org/abs/1703.00686}{Link}})\\
\cmidrule{3-5}          &       & Sun et al. \cite{sun2021fine}: \textcolor{gray}{Lightweight CNN with an attention mechanism (SENet) and joint learning for FGVC} & 2020  & Car-159: AP= 85.86\%  ({\small \href{https://link.springer.com/article/10.1007/s11042-020-09171-3}{Link}})\\
\cmidrule{3-5}          &       & Boukerche \& Ma \cite{boukerche2021novel}: \textcolor{gray}{A Lightweight Recurrent Attention Unit (LRAU) to enhance CNNs for VMMR} & 2022  & Stanford Cars: ACC= 92.64\% ({\small \href{https://ieeexplore.ieee.org/document/9660781}{Link}})\\
\cmidrule{3-5}          &       & Lu et al. \cite{lu2023amlnet}: \textcolor{gray}{An attention multibranch loss CNN that learns  regional features and their interactions} & 2023  & Stanford Cars: ACC= 94.18\% ({\small \href{https://ieeexplore.ieee.org/document/10234111}{Link}})\\
\midrule  
\multicolumn{1}{c|}{\multirow{7}[10]{*}{VLR}} & \multicolumn{1}{c|}{\multirow{2}[4]{*}{\shortstack{Handcrafted\\ features}}} & Chen et al. \cite{chen2016vehicle}: \textcolor{gray}{Symmetrical SURF for vehicle detection\& sparsity-based classification for model recognition} & 2016  & XMU: Accuracy=99.71\% \\
\cmidrule{3-5}          &       & Yu et al. \cite{yu2018vehicle}: \textcolor{gray}{Overlapping Enhanced Patterns of Oriented Edge Magnitudes with CRC for vehicle logo recognition} & 2018  & HFUT-VL: Accuracy=99.1\% \\
\cmidrule{2-5}          & \multicolumn{1}{c|}{\multirow{5}[1]{*}{\shortstack{Deep\\learning}}} & Huang et al. \cite{huang2015vehicle}: \textcolor{gray}{CNN for VMR that achieves robustness to poor  conditions without  precise logo segmentation} & 2015  & XMU: Accuracy=99.07\%  ({\small \href{https://ieeexplore.ieee.org/document/7031929}{Link}})\\
\cmidrule{3-5}          &       & Soon et al. \cite{soon2018hyper}: \textcolor{gray}{Particle swarm optimization  to automatically search and optimize CNN architecture for VLR} & 2018  & XMU: Accuracy=99.53\% ({\small \href{https://onlinelibrary.wiley.com/doi/10.1049/iet-its.2018.5127}{Link}})\\
\cmidrule{3-5}          &       & Li et al. \cite{li2024new}: \textcolor{gray}{Fine-tuned Swin Transformer for real-time Vehicle Logo Recognition} & 2024 & HFUT-VL1: Accuracy=99.28\% ({\small \href{https://arxiv.org/abs/2401.1545810.1049/iet-its.2018.5127}{Link}})\\
\midrule  
\multicolumn{1}{c|}{\multirow{10}[25]{*}{\shortstack{Vehicle\\ Re-ID}}} & \multicolumn{1}{c|}{\multirow{2}[4]{*}{\shortstack{Global\\ feature}}} & Li et al. \cite{li2017deep}: \textcolor{gray}{Deep joint discriminative learning model using a deep CNN to extract discriminative representations} & 2017  & Vehicle ID: CMC@1(small): 72.3\% ({\small \href{https://ieeexplore.ieee.org/document/8296310}{Link}})\\
\cmidrule{3-5}          &       & Zhang et al. \cite{zhang2017improving}: \textcolor{gray}{Improved triplet-wise CNN training for vehicle re-identification} & 2017  & Vehicle ID: CMC@1(small): 69.9\% ({\small \href{https://ieeexplore.ieee.org/document/8019491}{Link}})\\
\cmidrule{2-5}          & \multicolumn{1}{c|}{\multirow{4}[8]{*}{Local feature}} & Liu et al. \cite{liu2018ram}: \textcolor{gray}{Region-Aware deep Model (RAM) for extracting discriminative local features in vehicle Re-ID} & 2018  & \shortstack[l]{Vehicle ID: CMC@1(small): 75.2\%}({\small \href{https://ieeexplore.ieee.org/document/8486589}{Link}}) \\
\cmidrule{3-5}          &       & Huang et al. \cite{huang2023coarse}: \textcolor{gray}{Lightweight coarse-to-fine sparse self-attention for vehicle Re-ID} & 2023  & VeRi: mAP=78.5\% ({\small \href{https://www.sciencedirect.com/science/article/pii/S0950705123002769}{Link}})  \\
\cmidrule{3-5}          &       & Lian et al. \cite{lian2023multi}: \textcolor{gray}{Multi-branch network for extracting subtle distinguishing ReID features} & 2023  & \shortstack[l]{VeRi: mAP=83.4\%} ({\small \href{https://ieeexplore.ieee.org/document/10287201}{Link}}) \\
\cmidrule{3-5}          &       & Shen et al. \cite{shen2023git}: \textcolor{gray}{Cooperative graph (local) \& transformer (global) for robust ReID features} & 2023  & VeRi: mAP=80.3\%\% ({\small \href{https://ieeexplore.ieee.org/document/10287201}{Link}})\\
\cmidrule{2-5}          & \multicolumn{1}{c|}{\multirow{3}[5]{*}{\shortstack{Auxiliary\\information}}} & Chu et al. \cite{chu2019vehicle}: \textcolor{gray}{Viewpoint-Aware Network (VANet) that addresses the challenge of extreme viewpoint variation} & 2019  & \shortstack[l]{VeRi: mAP= 66.34\%} ({\small \href{https://ieeexplore.ieee.org/document/9010988}{Link}})\\
\cmidrule{3-5}          &       & Khorramshahi et al. \cite{khorramshahi2019dual}: \textcolor{gray}{Dual paths—global for macro, part-attention for local ReID features} & 2019  & \shortstack[l]{VeRi: mAP= 61.18\%}  ({\small \href{https://ieeexplore.ieee.org/document/9010356}{Link}})\\
\cmidrule{3-5}          &       & Quispe et al. \cite{quispe2021attributenet}: \textcolor{gray}{Distills color attributes, using Amelioration Constraint, for more discriminative ReID features} & 2021  &\shortstack[l]{ VeRi: mAP= 81.2\%} ({\small \href{https://www.sciencedirect.com/science/article/abs/pii/S0925231221013370}{Link}})\\
\cmidrule{3-5}          &       & Yu et al. \cite{yu2023semantic}: \textcolor{gray}{Multi-stage transformer coupling patches with learned, grouped semantics for ReID} & 2023  & \shortstack[l]{VeRi: mAP= 80.7\%} ({\small \href{https://ieeexplore.ieee.org/document/10081216}{Link}})\\
\midrule  
\multicolumn{1}{c|}{\multirow{5}[10]{*}{SOT}} & \multicolumn{1}{c|}{
{Filter}
}   & MEGTCF \cite{ma2022correlation}: \textcolor{gray}{Game theory orchestrates diverse (handcrafted/deep) experts for correlation filter tracking} & 2022  & OTB2015: SR=0.849 ({\small \href{https://ieeexplore.ieee.org/document/9783056/}{Link}})\\
\cmidrule{2-5}          & \multicolumn{1}{c|}{\multirow{3}[6]{*}{\shortstack{Siamese\\ network}}} & SiameseRPN \cite{li2018high}: \textcolor{gray}{Siamese region proposal network with classification and regression branches} & 2018  & OTB2015: SR=0.816 ({\small \href{https://arxiv.org/abs/2003.06761}{Link}})\\
\cmidrule{3-5}          &       & SiamBAN \cite{chen2020siamese}: \textcolor{gray}{Box adaptive network with anchor-free Siamese tracking framework} & 2020  & VOT2019: EAO=0.327 ({\small \href{https://arxiv.org/abs/2003.06761}{Link}})\\
\cmidrule{3-5}          &       & SiamDMU \cite{liu2024siamdmu}: \textcolor{gray}{Siamese tracker enhancing template updates via long-term motion \& semantic info.} & 2024  & VOT2018: EAO=0.427 ({\small \href{https://ieeexplore.ieee.org/document/10416263}{Link}})\\
\midrule  
\multicolumn{1}{c|}{\multirow{6}[11]{*}{MOT}} & \multicolumn{1}{c|}{\multirow{3}[6]{*}{SDT}} & DeepSORT \cite{wojke2017simple}: \textcolor{gray}{Deep appearance features with Kalman filtering for tracking-by-detection} & 2017  & MOT16: MOTA=61.4\% ({\small \href{https://arxiv.org/abs/1703.07402}{Link}})\\
\cmidrule{3-5}          &       & BYTETrack \cite{zhang2022bytetrack}: \textcolor{gray}{High-performance tracking with detection score association mechanism} & 2022  & MOT17: MOTA=78.6\% ({\small \href{https://arxiv.org/abs/2110.06864}{Link}})\\
\cmidrule{3-5}          &       & SMILEtrack \cite{wang2024smiletrack}: \textcolor{gray}{Occlusion-aware MOT by a siamese network-based similarity learning module } & 2024  & MOT17: MOTA=81.1\% ({\small \href{https://arxiv.org/abs/2211.08824}{Link}})\\
\cmidrule{2-5}          & \multicolumn{1}{c|}{\multirow{3}[5]{*}{JDE}} & FairMOT \cite{zhang2021fairmot}: \textcolor{gray}{Joint detection and embedding framework for balanced tracking performance} & 2021  & MOT16: MOTA=73.7\% ({\small \href{https://arxiv.org/abs/2004.01888}{Link}})\\
\cmidrule{3-5}          &       & TrackFormer \cite{meinhardt2022trackformer}: \textcolor{gray}{End-to-end trainable MOT approach based on an encoder-decoder Transformer architecture} & 2022  & MOT17: MOTA=74.1\% ({\small \href{https://arxiv.org/abs/2101.02702}{Link}})\\
\cmidrule{3-5}          &       & MeMOTR \cite{gao2023memotr}: \textcolor{gray}{Long-term memory-augmented Transformer for MOT using a custom memory-attention layer} & 2023  & MOT17: MOTA=72.8\% ({\small \href{https://arxiv.org/abs/2307.15700}{Link}})\\ \midrule

\end{longtable}
\end{minipage}

\textit{Vehicle keypoint-based} methods (Figure \ref{fig:highlevel_class}(b)) offer greater robustness in complex environments. Methods like AutoCalib \cite{bhardwaj2018autocalib} and approaches combining landmarks with vehicle classification \cite{bartl2020planecalib,bartl2021automatic} perform well even in irregular settings. However, they demand higher computational resources and depend heavily on accurate vehicle detection. Their calibration accuracy varies with traffic density—sparse conditions provide insufficient calibration data while congestion introduces keypoint occlusion challenges.

\subsubsection{Speed estimation}

Speed estimation in TSS calculates vehicle traveling speeds through video sequence analysis, primarily using two approaches: \textit{virtual section-based methods} and \textit{homography transformation-based} methods.

\begin{figure}[!htb]
  \centering
  \includegraphics[width=\textwidth]{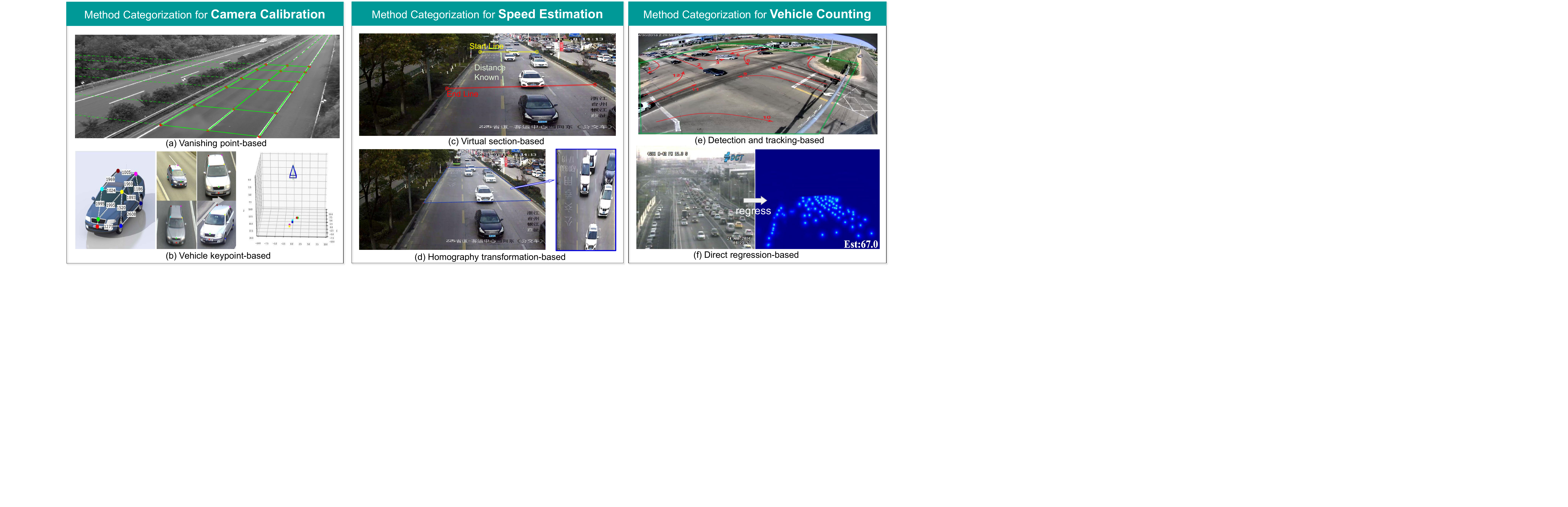}
  \caption{Schematic diagram of (a) vanishing point-based camera calibration methods; (b) vehicle keypoint-based camera calibration methods; (c) virtual section-based speed estimation methods; (b) homography transformation-based speed estimation methods; (e) detection and tracking-based vehicle counting methods and (f) direct regression-based  vehicle counting methods.}
  \label{fig:highlevel_class}
\end{figure}

\textit{Virtual section-based} methods, shown in Figure \ref{fig:highlevel_class}(c), use predefined virtual detection lines or areas to calculate speed by measuring vehicles' passage time through known distances \cite{celik2009solar,pornpanomchai2009vehicle,anandhalli2022image,ashraf2023hvd}. These methods have evolved from simple fixed-line implementations to more adaptive approaches, yet remain challenged by their fundamental reliance on manual calibration and sensitivity to camera positioning.

\textit{Homography transformation-based} methods, illustrated in Figure \ref{fig:highlevel_class}(d), transform image coordinates to real-world coordinates using homography matrices \cite{huang2018traffic,bell2020accurate,liu2020vision,yohannes2023improved,lashkov2023edge}. The evolution of these methods has primarily addressed three key challenges: reducing perspective distortion errors, minimizing manual calibration requirements, and improving computational efficiency while maintaining accuracy.

Despite numerous publications, a critical gap exists in quantitative performance comparison between methods. Recent surveys \cite{fernandez2021vision} attribute this to: lack of standardized datasets with reliable ground truth, inconsistent evaluation metrics, and the complexity of creating labeled benchmarking data. While reported error metrics suggest improvement from early implementations (3-7 km/h errors) to recent methods (below 2-3 km/h), these claims cannot be directly compared due to different testing conditions. Future research needs standardized benchmarking across diverse scenarios to enable rigorous assessment of methodological advantages.

\subsubsection{Vehicle counting}

Vehicle counting in TSS employs two complementary approaches designed for fundamentally different traffic scenarios: \textit{detection and tracking-based methods} for dynamic analysis and \textit{direct regression-based methods} for static estimation.

\textit{Detection and tracking-based} methods, shown in Figure \ref{fig:highlevel_class}(e), extract vehicle trajectories and implement counting rules using detection and tracking. Dai et al. \cite{dai2019video} combined YOLO v3 with KCF for multi-directional counting, while Song et al. \cite{song2019vision} developed YOLO v3+ORB for freeway analysis. Liu et al. \cite{liu2020vision} created a lane-specific method using YOLOv2 and Kalman filtering, and later \cite{liu2020robust} introduced a DTC framework for the AICity 2020 challenge. Majumder et al. \cite{majumder2023automated} implemented bidirectional counting through intersection tracking. However, these methods face several critical limitations: \textit{(1) accuracy dependency} on the performance of underlying detectors and trackers, where detection failures directly propagate to counting errors; \textit{(2) computational complexity} from the sequential detection-tracking pipeline, making real-time processing challenging for high-resolution multi-camera systems; \textit{(3) association failures} in dense traffic scenarios where target identity switches can cause significant counting discrepancies; and \textit{(4) environmental sensitivity} to occlusions, lighting variations, and weather conditions that degrade both detection and tracking performance.

\textit{Direct regression-based} methods, depicted in Figure \ref{fig:highlevel_class}(f), inspired by crowd counting \cite{khan2023revisiting}, use end-to-end neural networks for direct vehicle counting. Oñoro-Rubio et al. \cite{onoro2016towards} developed CCNN and Hydra CNN models, while Zhang et al. \cite{zhang2017fcn} introduced FCN-rLSTM combining CNN with LSTM. Yang et al. \cite{yang2021fast} proposed a TSI approach, and Guo et al. \cite{guo2023scale} created SRRNet with SLA and ORR features. Despite their efficiency advantages, these methods suffer from significant limitations: \textit{(1) dataset dependency} requiring extensive training data representative of target deployment scenarios; \textit{(2) limited generalization} across different camera viewpoints, traffic densities, and environmental conditions; \textit{(3) density estimation inaccuracy} in highly congested scenes where vehicle overlap complicates counting; and \textit{(4) operational constraints} including inability to provide lane-specific traffic volume, directional flow analysis, or individual vehicle trajectories essential for comprehensive traffic analysis.

These two approaches serve distinct traffic monitoring needs rather than representing evolutionary improvements of the same technique. The selection between them should be guided by specific application requirements: dynamic trajectory analysis versus static density estimation, with careful consideration of their respective accuracy dependencies and operational limitations in target deployment environments.
\begin{figure}[!tb]
  \centering
  \includegraphics[width=\textwidth]{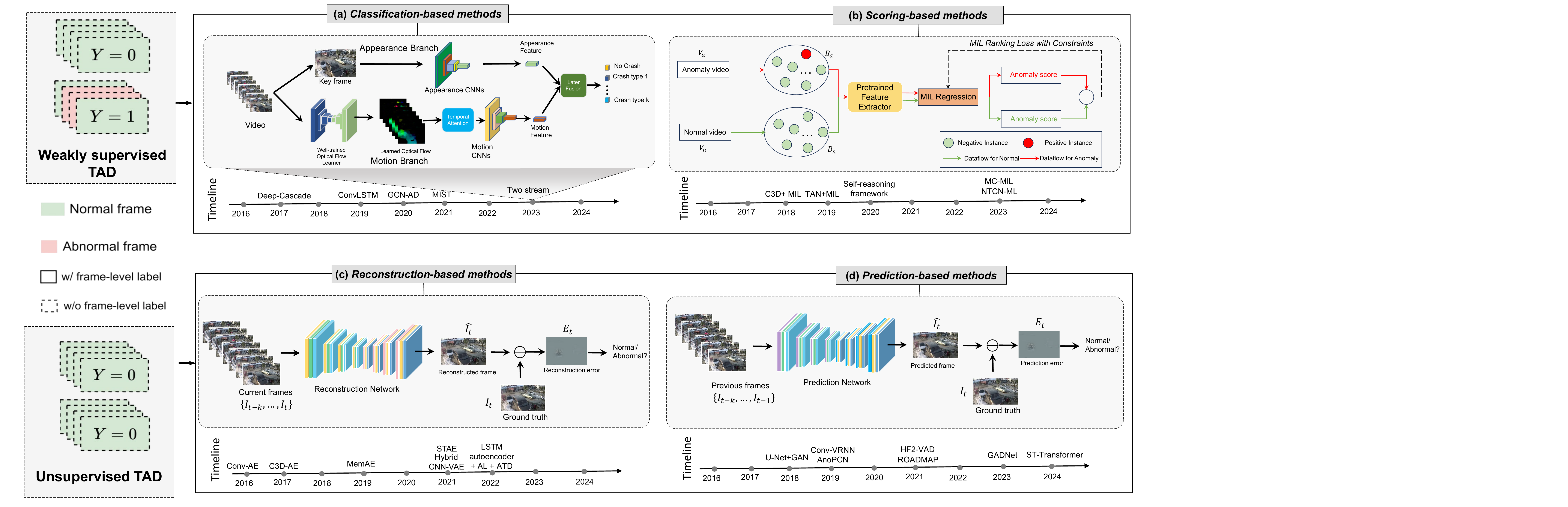}
  \caption{  Categorization and development timeline of current traffic anomaly detection (TAD), which includes weakly supervised and unsupervised learning approaches. Weakly supervised methods can be divided into classification-based and scoring-based categories, whereas unsupervised learning methods comprise reconstruction-based and prediction-based approaches.}
  \label{fig11}
\end{figure}
\subsection{Traffic Anomaly Detection}

Traffic anomaly detection in TSS identifies behaviors deviating from normal patterns, including accidents, violations, and unusual congestion. According to \cite{liu2024generalized}, approaches are categorized into \textit{weakly supervised} and \textit{unsupervised learning} paradigms, as shown in Figure \ref{fig11}. Weakly supervised learning uses video-level labels indicating anomaly presence, while unsupervised learning detects anomalies without labeled data.

\subsubsection{Weakly supervised traffic anomaly detection}
Weakly Supervised Traffic Anomaly Detection (WSTAD) utilizes video-level labels and comprises two main approaches: \textit{classification-based} and \textit{scoring-based} methods.

\textit{Classification-based} methods directly classify videos as normal or anomalous. Sabokrou et al. \cite{sabokrou2017deep} developed a cubic-patch-based approach with cascaded classifiers. Batanina et al. \cite{batanina2019domain} created a 3D CNN for accident detection with dual classification heads. Lu et al. \cite{lu2020new} integrated ResNet with attention modules for crash detection. Zhong et al. \cite{zhong2019graph} employed graph convolutional networks, while Feng et al. \cite{feng2021mist} introduced the MIST framework. Zhou et al. \cite{zhou2023appearance} developed an appearance-motion network for crash detection (Figure \ref{fig11} (a)). Yu et al. \cite{yu2024fine} proposed a transformer-based framework with the FAD database. These methods offer straightforward implementation with good detection accuracy, evolving from limited feature representation in early work to sophisticated architectures addressing temporal modeling, but continue to struggle with anomaly localization and generalization to novel anomaly types.

\textit{Scoring-based} methods assign anomaly scores using Multiple Instance Learning (MIL) ranking frameworks (Figure \ref{fig11}(b)). Sultani et al. \cite{sultani2018real} pioneered deep multiple instance ranking. Zhu et al. \cite{zhu2019motion} enhanced MIL with attention mechanisms, while Zaheer et al. \cite{zaheer2020self} developed self-reasoning through clustering. Shao et al. \cite{shao2023video} introduced NTCN-ML, and Pereira et al. \cite{pereira2024mc} proposed MC-MIL for multi-camera scenarios. These approaches excel at temporal localization and handling imbalanced datasets. Recent developments have focused on three primary improvement directions: enhanced feature extraction through attention mechanisms \cite{zhu2019motion}, label denoising techniques to handle noisy video-level annotations\cite{zhong2019graph}, and objective function refinements for better ranking performance \cite{liu2022abnormal}. However, these methods typically require more complex training procedures and remain computationally expensive due to their dependency on pre-trained feature extractors.

Despite advances, WSTAD methods face three key limitations: coarse-grained video-level labels that miss subtle anomalies, limited generalization to novel anomaly types, and poor performance on imbalanced datasets where anomalous events are rare.

\subsubsection{Unsupervised traffic anomaly detection}

\begin{figure}[!tb]
  \centering
  \includegraphics[width=0.7\textwidth]{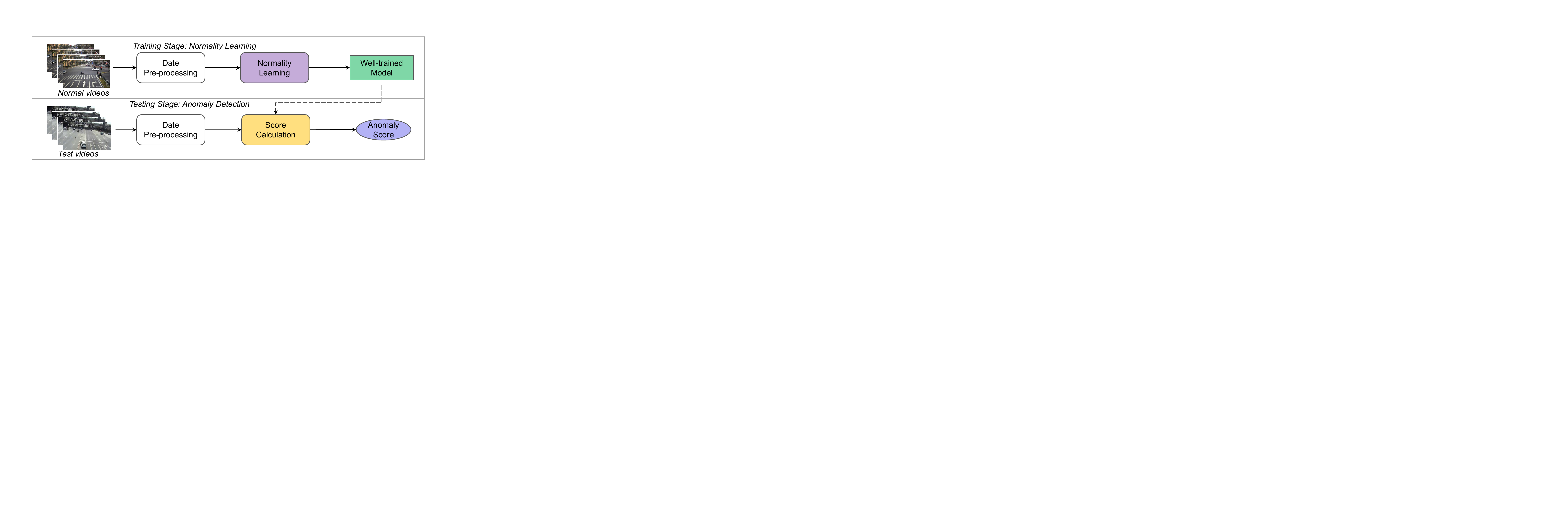}
  \caption{ Two-stage nature of Unsupervised Traffic Anomaly Detection (UTAD), which learns normal patterns at training stage and detects anomalies at testing stage.}
  \label{fig12}
\end{figure}

Unsupervised Traffic Anomaly Detection (UTAD) identifies anomalies without labeled data, particularly valuable for undefined anomalies or scenarios lacking labeled data. As shown in Figure \ref{fig12}, UTAD methods follow a two-stage process (learning normal patterns, then detecting anomalies) and divide into \textit{reconstruction-based} and \textit{prediction-based} methods.

\textit{Reconstruction-based} methods \cite{gong2019memorizing,hasan2016learning,deepak2021residual,santhosh2021vehicular} identify anomalies through reconstruction errors using autoencoder architectures. Hasan et al. \cite{hasan2016learning} developed autoencoder approaches using both handcrafted features and end-to-end learning. Gong et al. \cite{gong2019memorizing} introduced MemAE with a memory module for normal patterns. Deepak et al. \cite{deepak2021residual} proposed residual STAE for pattern reconstruction. For trajectory analysis, Santhosh et al. \cite{santhosh2021vehicular} developed a CNN-VAE architecture, while Zhou et al. \cite{zhou2022vision} created an LSTM autoencoder with adversarial learning. These methods benefit from requiring no labeled anomaly data and effectively capturing spatial patterns, progressing from basic autoencoders to memory-augmented and adversarial approaches, though they still struggle with subtle anomalies that can be reconstructed accurately.

\textit{Prediction-based} methods \cite{liu2018future,liu2021hybrid,wang2021robust,tran2024transformer} detect anomalies by comparing predicted patterns with actual observations. Liu et al. \cite{liu2018future} pioneered future frame prediction with spatial-temporal constraints, later enhanced by Liu et al. \cite{liu2021hybrid} with HF2-VAD. Wang et al. \cite{wang2021robust} proposed multi-path ConvGRU for various scales, while Tran et al. \cite{tran2024transformer} introduced a transformer-based approach for complex scenes. These approaches demonstrate superior performance in capturing temporal dynamics, with advancement from early frame prediction to multi-scale and transformer-based models addressing limitations in handling complex scenes, though they remain computationally intensive and may struggle with purely appearance-based anomalies.

While UTAD methods show progress, they primarily struggle with their dependence on extensive normal video data. This limitation complicates the definition of normal behavior, affecting model adaptability in dynamic real-world traffic scenarios where normality patterns continuously evolve.

\subsection{Traffic Behavior Understanding}

Traffic behavior understanding in TSS analyzes traffic participants' movements and interactions, focusing on \textit{\textbf{recognition}} and \textit{\textbf{prediction}} of behavioral patterns. Due to distinct characteristics between vehicles and vulnerable road users (pedestrians and cyclists), the field divides into two domains: \textit{Vehicle Behavior Understanding} (VBU) and \textit{Vulnerable Road User Behavior Understanding} (VRBU), as shown in Figure \ref{fig13}.

While TSS-specific research remains limited, methodologies from dashcam and UAV perspectives can be adapted to TSS applications. This section reviews traffic behavior understanding approaches across multiple viewpoints to derive TSS-applicable insights.

\subsubsection{Vehicle Behavior Understanding}

\begin{figure}[!tb]
  \centering
  \includegraphics[width=0.9\textwidth]{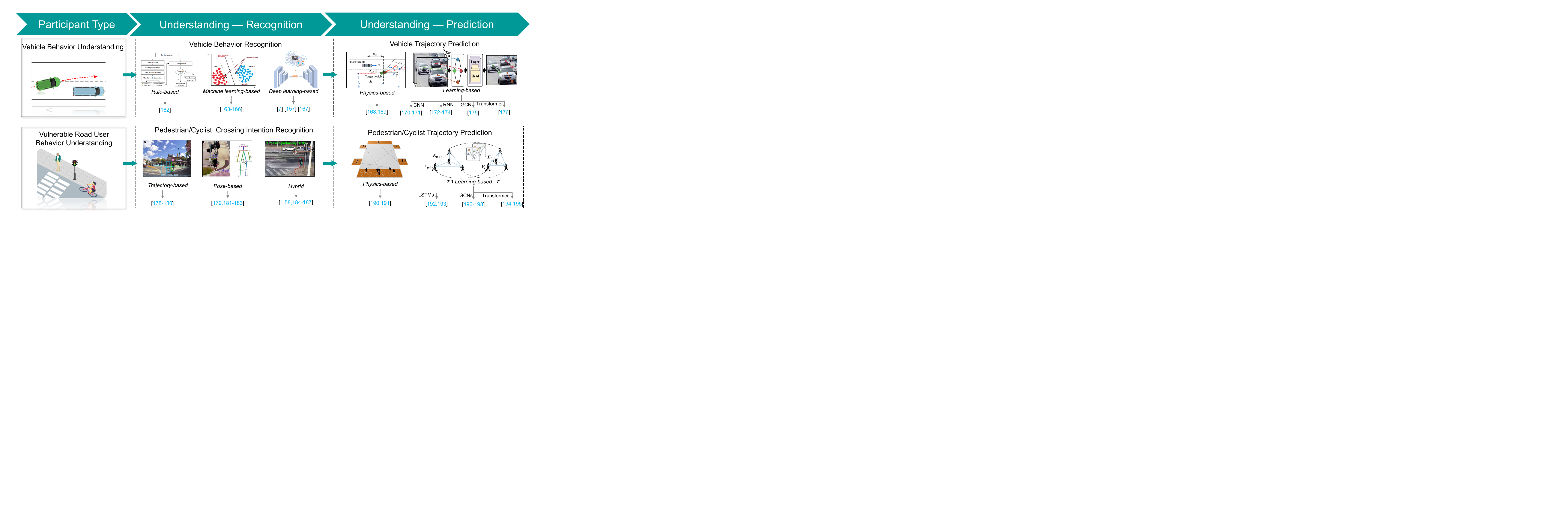}
  \caption{ Categorization and literature of Vehicle Behavior Understanding (VBU) and Vulnerable Road User Behavior Understanding (VRBU).}
  \label{fig13}
\end{figure}

Vehicle Behavior Understanding (VBU), as shown in Figure \ref{fig13}, aims to \textit{\textbf{recognize}} and \textit{\textbf{predict}} complex vehicle actions including lane changing, turning, speed variations, and traffic violations. 

Vehicle behavior recognition primarily relies on trajectory analysis through traditional and deep learning methods. Traditional approaches employ various techniques including decision rules \cite{song2014vehicle}, genetic algorithms \cite{uy2015machine}, SVM \cite{aoude2012driver}, ensemble KNN \cite{zhang2021ensemble}, and LGBM \cite{xu2023integrating}. While effective, these methods require extensive feature engineering. Deep learning approaches \cite{santhosh2020anomaly,zhou2022vision,haghighat2023computer} demonstrate superior performance in complex scenarios, with Santhosh \cite{santhosh2020anomaly} developing a CNN-VAE architecture and Haghighat \cite{haghighat2023computer} achieving high accuracy in violation detection, though requiring substantial labeled data. The evolution reveals a paradigm shift from rule-based feature engineering to end-to-end learning. Traditional methods offer interpretability and high deployment feasibility in resource-constrained environments due to their computational efficiency, but their real-world applicability is limited by poor adaptability to diverse conditions and environmental vulnerability. Deep learning approaches excel in generalization and robustness, offering better potential for complex real-world scenarios, but their high data dependency and computational intensity currently hinder widespread deployment feasibility.

With the advancement of autonomous driving and V2X technologies, vehicle trajectory prediction has become increasingly crucial for safety warnings and decision-making. These predictions analyze current movement patterns and environmental context to forecast future trajectories. Methods fall into two categories: \textit{physics-based} and \textit{learning-based} models. \textit{Physics-based} models \cite{xie2017vehicle,anderson2021kinematic} use kinematic models, Gaussian processes, and Bayesian networks, offering interpretability but limited effectiveness in complex scenarios. \textit{Learning-based} models leverage CNNs \cite{nikhil2018convolutional,katariya2022deeptrack}, RNNs \cite{lin2021vehicle,zhang2022d2,guo2022vehicle}, GCNs \cite{ren2024emsin}, and Transformers \cite{pazho2024vt}. Notable examples include Yuan et al.'s \cite{yuan2023temporal} TMMOE model and Pazho et al.'s \cite{pazho2024vt} VT-Former for surveillance scenarios. 

Physics-based models provide theoretical guarantees and interpretability, making them suitable for scenarios where these are paramount and enhancing their deployment feasibility in safety-critical but simpler contexts; however, their simplified assumptions limit their real-world applicability in complex, dynamic environments. Learning-based models demonstrate superior real-world applicability in modeling complex interactions, yet their significant data requirements and generalization challenges across varied scenarios can impede straightforward deployment feasibility. The field progresses toward \textit{hybrid architectures}, \textit{attention mechanisms}, and \textit{uncertainty-aware predictions} to improve overall reliability and make learning-based approaches more robust for real-world deployment.

\subsubsection{Vulnerable Road User Behavior Understanding}

Vulnerable Road User Behavior Understanding primarily focuses on Crossing Intention Recognition (CIR) and Trajectory Prediction (TP). These areas are crucial for traffic safety, as pedestrian and cyclist behavior patterns strongly correlate with accident rates. Crossing intentions are categorized into Crossing (C) and Non-Crossing (NC), while trajectory prediction forecasts future positions over time. As shown in Figure \ref{fig13}, current methodologies classify into three categories: \textit{Trajectory-based} \cite{goldhammer2019intentions,saleh2018intent,xu2022remember}, \textit{Pose-based} \cite{saleh2018intent,zhang2021pedestrian,fang2019intention,xu2021action}, and \textit{Hybrid CIR} models \cite{zhou2023pedestrian,quispe2021attributenet,rasouli2017they,gesnouin2021trouspi,kotseruba2021benchmark,azarmi2024pip}.

Early \textit{Trajectory-based} CIR models \cite{goldhammer2019intentions,saleh2018intent,xu2022remember} analyzed historical movement patterns, but showed insufficient prediction accuracy \cite{zhang2023pedestrian}. This led to \textit{pose-based} models \cite{saleh2018intent,zhang2021pedestrian,fang2019intention,xu2021action}, incorporating body orientation and gestural signals. Notable examples include Xu et al.'s \cite{xu2021action} work combining 3D pose estimation with adaptive graph networks, and Zhang et al.'s \cite{zhang2021pedestrian} approach using pose estimation for red-light crossing behavior prediction.

Current \textit{hybrid} models \cite{zhou2023pedestrian,quispe2021attributenet,rasouli2017they,gesnouin2021trouspi,kotseruba2021benchmark,azarmi2024pip} integrate trajectories, poses, and environmental context, showing superior performance in complex scenarios. Key developments include the Dual-Channel Network \cite{yang2023dpcian} for modeling poses and environmental interactions, PIP-Net \cite{azarmi2024pip} integrating multiple input types, and Zhou et al.'s \cite{zhou2023pedestrian} pedestrian-centric approach. While more accurate, these methods require higher computational resources.

CIR evolution demonstrates progressive multimodal integration. Trajectory-based approaches offer high deployment feasibility due to their efficiency, but their real-world applicability is constrained by temporal limitations and an inability to capture subtle behavioral cues crucial for accurate intention recognition. Pose-based methods provide finer-grained analysis, improving real-world applicability for capturing subtle intentions, but their computational complexity can challenge deployment feasibility in real-time systems. Hybrid models achieve superior performance through comprehensive context modeling and represent the current mainstream, offering the best real-world applicability in complex scenes; however, they still face increased computational demands and multimodal fusion complexity, which are key hurdles for widespread deployment feasibility.

Trajectory prediction approaches divide into \textit{physics-based} \cite{alahi2014socially,ningbo2017simulation} and \textit{learning-based} models \cite{alahi2016social,gupta2018social,shi2023trajectory,zhang2023forceformer}. \textit{Physics-based} models utilize hand-crafted features and social force models to quantify interactions. \textit{Learning-based} models have evolved along three paths: LSTMs \cite{alahi2016social,gupta2018social} for processing sequential data and capturing temporal dependencies, GCNs \cite{shi2021sgcn,lian2023ptp,lv2023ssagcn} for modeling spatial relationships, and Transformers \cite{shi2023trajectory,zhang2023forceformer} for handling complex interactions in crowded scenarios.

The evolution from physics-based to learning-based models reflects increasing sophistication in social interaction modeling. Physics-based approaches provide excellent deployment feasibility due to computational efficiency and interpretability, making them suitable for simpler, well-defined scenarios, but their real-world applicability is limited in complex, crowded social interactions. Learning-based progression (LSTMs, GCNs, Transformers) demonstrates enhanced handling of complex dependencies, boosting their potential real-world applicability across diverse and crowded scenarios. However, the escalating computational complexity and significant training challenges associated with these advanced models pose substantial hurdles to their immediate deployment feasibility without further optimization and research into model efficiency.

\begin{table*}[htbp]
\makeatletter\setlength{\@fptop}{0pt}\makeatother
\hspace*{-2cm} % Adjust this negative space as needed for your page layout
\footnotesize
\begin{minipage}{\dimexpr\textwidth+3cm\relax} % Adjust total width as needed
  \centering
    \renewcommand\arraystretch{0.8} % Kept as per your original for this table
  \caption{Overview of common datasets for high-level perception tasks in TSS (including traffic parameter estimation, traffic anomaly detection, and traffic behavior understanding), with sensor details and TSS relevance. Size units: \textcolor{blue}{I} (Images), \textcolor{orange}{V} (Videos), \textcolor{green}{S} (Samples).}
    % Adjusted column widths: m{1.5cm} for Task, m{1.7cm} for Sub-Task, m{2.2cm} for Dataset, m{0.7cm} for Year, m{1.5cm} for Size, m{1.8cm} for Sensor Platform, m{5.6cm} for Key Characteristics, m{2.5cm} for Link
    \begin{tabular}{>{\centering\arraybackslash}m{1.4cm}|>{\centering\arraybackslash}m{1.2cm}|>{\centering\arraybackslash}m{2cm}|>{\centering\arraybackslash}m{0.5cm}|>{\centering\arraybackslash}m{1.5cm}|>{\centering\arraybackslash}m{1.8cm}|>{\centering\arraybackslash}m{8.6cm}|>{\arraybackslash}m{0.7cm}}
    \toprule
    \toprule
    \multicolumn{1}{c|}{{\textbf{Task}}} & \multicolumn{1}{c|}{{\textbf{Sub-Task}}} & {\textbf{Dataset}} & \multicolumn{1}{c|}{{\textbf{Year}}} & \textbf{Size \textbf{(\textcolor{blue}{I};\textbf{\textcolor{orange}{V};\textbf{\textcolor{green}{S})}}}} & {\textbf{Sensor Platform}} & {\textbf{Key Characteristics \& TSS Focus}} & {\textbf{Link}}  \\
    \midrule
    \midrule
    \multirow{8}[16]{*}{\shortstack{Traffic \\ Parameter \\Estimation}} & \multirow{4}[8]{*}{\shortstack{Speed \\Evaluation}} & AI City 2018 Track 1~\cite{naphade20182018} & 2018 & 142 \textcolor{orange}{V} & Surveillance & Traffic footage with vehicle-level annotations from 27 different camera views; Supports multi-camera traffic speed estimation and flow monitoring in signalized intersections. & {\small \href{https://www.aicitychallenge.org/2018-ai-city-challenge/}{Link}} \\
\cmidrule{3-8}          &       & BrnoCompSpeed~\cite{sochor2018boxcars} & 2018 & 18 \textcolor{orange}{V} & Surveillance & Highway surveillance with precise LIDAR-based speed measurements yet limited lighting/weather conditions; Enables development of  speed measurement systems from different viewpoints. & {\small \href{https://github.com/JakubSochor/BrnoCompSpeed}{Link}} \\
\cmidrule{3-8}          &       & UTFPR~\cite{luvizon2014vehicle} & 2014 & 20 \textcolor{orange}{V} & Surveillance &  Low-cost sensor videos (varied conditions) with precise speed \& license plate ground truth; Supports monocular speed estimation for cost-effective traffic monitoring. & N/A \\
\cmidrule{3-8}          &       & QMUL\cite{hospedales2012video} & 2012 & 1 \textcolor{orange}{V} & Surveillance & Nearly one-hour continuous high-density traffic footage at a complex urban junction; Supports analysis of traffic patterns and anomaly detection at signalized intersections. & {\small \href{http://www.eecs.qmul.ac.uk/~sgg/QMUL_Junction_Datasets/Junction2/Junction2.html}{Link}} \\
\cmidrule{2-8}          & \multirow{4}[8]{*}{\shortstack{Vehicle \\Counting}} & Freeway-vehicle~\cite{song2019vision} & 2019 & 11,129 \textcolor{blue}{I} & Surveillance & Large-scale HD highway surveillance imagery with fully annotated vehicles in diverse scenes, featuring significant scale variations; Enables development and evaluation of vehicle detection/counting algorithms. & {\small \href{http://drive.google.com/open?id=1li858elZvUgss8rC_yDsb5bDfiRyhdrX}{Link}} \\
\cmidrule{3-8}          &       & AI City 2020 Track-1~\cite{chang2020ai} & 2020 & 31 \textcolor{orange}{V} & Surveillance & Multi-camera urban footage focused on counting vehicles (cars, trucks) executing pre-defined movements (turns, through traffic) at intersections; Supports detailed traffic demand analysis and optimization of signal timing for congestion mitigation. & {\small \href{https://www.aicitychallenge.org/2020-ai-city-challenge/}{Link}} \\
\cmidrule{3-8}          &       & TRANCOS~\cite{guerrero2015extremely} & 2015 & 1,244 \textcolor{blue}{I} & Surveillance & Highly congested traffic images with over 46,000 annotated vehicles and perspective distortion; Supports density-based counting approaches for crowded traffic scenes with severe vehicle occlusion. & {\small \href{http://agamenon.tsc.uah.es/Personales/rlopez/data/trancos}{Link}} \\
\cmidrule{3-8}          &       & CARPK~\cite{hsieh2017drone} & 2017 & 1,448 \textcolor{blue}{I} & UAV & Aerial imagery of parking lots with nearly 90,000 annotated cars from various viewpoints; Facilitates development of automated parking management and occupancy monitoring applications. & {\small \href{https://lafi.github.io/LPN/}{Link}} \\
    \midrule
    \multirow{11}[14]{*}{\shortstack{Video \\Anomaly \\Detection}} & \multirow{4}[8]{*}{\shortstack{General-\\purpose}} & UCSD Ped1/2~\cite{li2013anomaly} & 2013 & 18,560 \textcolor{blue}{I} & Surveillance & Pedestrian walkways with anomalies defined as non-pedestrian entities or unusual movement patterns; Enables detection of safety-critical events in pedestrian-oriented areas within traffic environments. & {\small \href{http://www.svcl.ucsd.edu/projects/anomaly/dataset.html}{Link}} \\
\cmidrule{3-8}          &       & CUHK-Avenue~\cite{lu2013abnormal} & 2013 & 30,652 \textcolor{blue}{I} & \shortstack{Internet \& \\Surveillance} & Campus avenue scenarios with abnormal events including running, wrong direction movement, and abandonment; Supports security monitoring for detecting unusual human behaviors in public transit areas. & {\small \href{http://www.cse.cuhk.edu.hk/leojia/projects/detectabnormal/dataset.html}{Link}} \\
\cmidrule{3-8}          &       & Shanghai Tech~\cite{liu2018future} & 2018 & 300,308 \textcolor{blue}{I} & Surveillance & Large-scale complex scenes from 13 different campus locations with diverse viewing angles; Enables robust anomaly detection across varied urban environments with challenging real-world conditions. & {\small \href{https://svip-lab.github.io/dataset/campus_dataset.html}{Link}} \\
\cmidrule{3-8}          &       & UCF-Crime~\cite{sultani2018real} & 2018 & 13.7M \textcolor{blue}{I} & Surveillance & Large-scale surveillance dataset (1900 videos) covering 13 public safety anomalies with video-level annotations; Supports both general anomaly detection and specific incident recognition in surveillance settings. & {\small \href{https://webpages.uncc.edu/cchen62/dataset.html}{Link}} \\
\cmidrule{2-8}          & \multirow{7}[6]{*}{\shortstack{Traffic-\\Specific}} & CADP~\cite{shah2018cadp} & 2018 & 1,416 \textcolor{orange}{V} & Surveillance & YouTube-sourced traffic camera dataset  focused on complex accident scenes; Enables research in accident detection, prediction, and small-object detection challenges in traffic surveillance. &{\small \href{https://ankitshah009.github.io/accident_forecasting_traffic_camera}{Link}}  \\
\cmidrule{3-8}          &       & {CDD~\cite{zhou2023appearance}} &{2023} & {6,166 \textcolor{orange}{V}} & {Surveillance} & Large-scale traffic crash dataset with diverse crash types spanning single-vehicle, multi-vehicle, non-motorized, and pedestrian incidents; Enables real-time crash detection systems for immediate emergency response coordination. & {\small \href{https://github.com/vvgoder/Dataset_for_crashdetection}{Link}}  \\
\cmidrule{3-8}          &       & UIT-ADrone~\cite{tran2023uit} & 2023 & 206,194 \textcolor{blue}{I} & UAV & Drone-captured dataset  featuring 10 anomaly types at complex urban roundabouts in Vietnam; Enables development of aerial anomaly detection systems for challenging traffic environments. & {\small \href{https://uit-together.github.io/datasets/UIT-ADrone/}{Link}} \\
    \midrule
    \multirow{13}[18]{*}{\shortstack{Behavior \\Understanding}} & {\shortstack{Pedestrian \\Trajectory \\Prediction}} & \shortstack{ETH\cite{pellegrini2010improving}\\UCY~\cite{lerner2007crowds}} & \shortstack{2009\\2007} & {2,206 \textcolor{green}{S}} & {UAV} & Widely-used combined benchmark of bird's-eye view recordings with precise trajectory annotations across 5 diverse scenes; Standard evaluation platform for pedestrian trajectory prediction in varied urban environments. & \shortstack{{\small \href{https://data.vision.ee.ethz.ch/cvl/aem/ewap_dataset_full.tgz}{Link}} \\ {\small \href{https://graphics.cs.ucy.ac.cy/research/downloads/crowd-data}{Link}}} \\
\cmidrule{2-8}          & \multirow{4}[3]{*}{\shortstack{Pedestrian \\Intention\\ Recognition}} & {JAAD~\cite{rasouli2017they}} & {2017} & 2.8k \textcolor{green}{S} \& 82k \textcolor{blue}{I} & {Dashcam} & Urban driving dataset  with rich annotations for pedestrian-driver joint attention behaviors and interactions; Features detailed bounding boxes, demographic attributes, behavioral labels, and scene elements to support research in pedestrian detection and behavior prediction. & {\small \href{http://data.nvision2.eecs.yorku.ca/JAAD_dataset/}{Link}}   \\
\cmidrule{3-8}          &       & {PIE~\cite{rasouli2019pie}} & {2019} & 1.8k \textcolor{green}{S} \& 911k \textcolor{blue}{I}& {Dashcam} & Large-scale dataset specifically designed for pedestrian intention estimation and trajectory prediction in urban traffic environments; Facilitates predictive models for vulnerable road user safety at complex urban crossings. & {\small \href{http://data.nvision2.eecs.yorku.ca/PIE_dataset/}{Link}}   \\
\cmidrule{2-8}          & \multirow{3}[6]{*}{\shortstack{Vehicle \\Behavior\\ Recognition}} & NGSIM & 2007 & 1.75 hrs \textcolor{orange}{V} & UAV & High-resolution trajectory data of vehicles on US highways with lane-level precision; Supports microscopic traffic flow modeling and car-following behavior analysis for traffic simulation. & {\small \href{http://ngsim.fhwa.dot.gov}{Link}} \\
\cmidrule{3-8}          &       & HighD~\cite{krajewski2018highd} & 2018 & 16.5 hrs \textcolor{orange}{V} & UAV & Drone-recorded vehicle trajectories on German highways with 110,000+ vehicles; Enables analysis of naturalistic driving patterns and vehicle interactions for advanced traffic management. & {\small \href{https://levelxdata.com/highd-dataset/}{Link}} \\
\cmidrule{3-8}          &       & CitySim~\cite{zheng2024citysim} & 2022 & 19 hrs \textcolor{orange}{V} & UAV & Comprehensive drone-based dataset with trajectories from 1,140 minutes of video across 12 diverse road locations; Enables accurate traffic monitoring, safety analysis, and trajectory prediction for traffic surveillance systems. & {\small \href{https://github.com/ozheng1993/UCF-SST-CitySim-Dataset}{Link}} \\
\cmidrule{2-8}          & \multirow{4}[6]{*}{\shortstack{Vehicle\\Trajectory\\ Prediction}} & {Apolloscape~\cite{huang2018apolloscape}} & {2019} & 140k \textcolor{blue}{I} \& 73 \textcolor{orange}{V} & {Dashcam} & Multi-sensor urban driving data with 3D perception and diverse traffic participants; Supports motion prediction modeling in complex urban scenarios for proactive traffic management. & {\small \href{https://apolloscape.auto/}{Link}}   \\
\cmidrule{3-8}          &       & Lyft L5~\cite{houston2021one} & 2021 & 1k+ hrs \textcolor{orange}{V} & Dashcam & Large-scale autonomous driving dataset with semantic maps and multi-agent trajectories; Enables advanced motion forecasting for diverse road users in dynamic urban environments. & {\small \href{https://self-driving.lyft.com/level5/prediction/}{Link}} \\
\cmidrule{3-8}          &       & V2X-Seq~\cite{yu2023v2x} & 2023 & 200k+ \textcolor{orange}{V} & \shortstack{Dashcam \& \\Surveillance} & Vehicle-infrastructure cooperative sensing data with synchronized multi-view perspectives; Facilitates research on enhanced perception and prediction through V2X communication for intelligent traffic systems. & {\small \href{https://github.com/AIR-THU/DAIR-V2X-Seq}{Link}} \\
    \bottomrule
    \end{tabular}%
  \label{tab3}%
\end{minipage}
\end{table*}

\subsection{Performance evaluation}
This section first details the datasets and evaluation metrics used for these high-level perception tasks in TSS, including traffic parameter estimation, traffic anomaly detection and traffic behavior understanding. After that, the results of some representative approaches are presented. 

\subsubsection{Datasets for high-level perception}

In the field of traffic parameter estimation, representative datasets include AI City Challenge \cite{naphade20182018}, BrnoCompSpeed \cite{sochor2018boxcars}, UTFPR \cite{luvizon2014vehicle}, and QMUL\footnote{https://www.eecs.qmul.ac.uk/~sgg/QMUL\_Junction\_Datasets/Junction/Junction.html}  for speed evaluation, as well as Freeway-vehicle dataset \cite{song2019vision}, AI City 2020 Track-1 \cite{chang2020ai}, TRANCOS \cite{guerrero2015extremely} and CARPK \cite{hsieh2017drone} for vehicle counting.

In the field of video anomaly detection, representative general-purpose datasets include UCSD Ped1/Ped2 \cite{li2013anomaly}, CUHK-Avenue \cite{lu2013abnormal}, Shanghai Tech \cite{liu2018future}, and UCF-Crime \cite{sultani2018real}, which have been widely adopted for traffic anomaly detection despite their broader scope. For traffic-specific anomaly detection, specialized datasets have been developed, such as CADP \cite{shah2018cadp}, CDD \cite{zhou2023appearance}, and UIT-ADrone \cite{tran2023uit}.

In the field of traffic behavior understanding, representative datasets can be categorized by their specific focuses. For pedestrian behavior analysis, datasets include trajectory prediction-oriented ETH/UCY \cite{pellegrini2010improving,lerner2007crowds} and intention recognition-focused JAAD \cite{rasouli2017they} and PIE \cite{rasouli2019pie}. Vehicle behavior datasets comprise three categories: general behavior recognition datasets such as NGSIM \footnote{https://data.transportation.gov/Automobiles/Next-Generation-Simulation-NGSIM-Vehicle-Trajector/8ect-6jqj}, HighD \cite{krajewski2018highd}, and CitySim \cite{zheng2024citysim}, autonomous driving datasets including Apolloscape \cite{huang2018apolloscape} and Lyft L5 \cite{houston2021one}, and vehicle-infrastructure cooperative datasets like V2X-Seq \cite{yu2023v2x}. More detailed statistics is shown in Table \ref{tab3}.

\subsubsection{Metrics and performance evaluation}

For traffic parameter estimation, the performance of speed estimation is commonly evaluated using three primary metrics: Mean Absolute Error (MAE) expressed in km/h to measure average estimation error, Mean Square Error (MSE), and Root Mean Square Error (RMSE) \cite{pornpanomchai2009vehicle,anandhalli2022image,ashraf2023hvd}. As for vehicle counting, evaluation metrics vary by methodology: detection and tracking-based approaches commonly use Mean Percentage Error (MPE) and Mean Correct Rate (MCR) \cite{dai2019video,liu2020robust,majumder2023automated}, while regression-based methods prefer Mean Absolute Error (MAE) and Grid Average Mean Error (GAME) \cite{khan2023revisiting,onoro2016towards}. 

For traffic anomaly detection, which generally operates as a binary classification task \cite{liu2024generalized}, the primary evaluation metrics include the Receiver Operating Characteristic (ROC) curve and its Area Under the Curve (AUC). Additionally, due to the inherent class imbalance in anomaly detection scenarios, F1-Score, which combines precision and recall, is commonly used alongside traditional accuracy measurements \cite{zhou2023appearance}.

Traffic behavior understanding tasks employ different evaluation metrics based on their specific objectives. For behavior recognition and intention prediction, which are classification tasks, common metrics include Accuracy, F1-score, Precision, Recall, and Average Precision (AP) \cite{zhou2023pedestrian}. For trajectory prediction of vehicles and vulnerable road users, which is treated as a regression problem, the primary metrics are Average Displacement Error (ADE) and Final Displacement Error (FDE) \cite{alahi2016social,gupta2018social}, measuring the average and final position errors between predicted and ground truth trajectories. Additional metrics such as RMSE 
\cite{yuan2023temporal}, collision rate \cite{liu2020col,kothari2021human}, and negative log-likelihood \cite{kothari2021human} are also employed in specific studies. 

Table \ref{tab4} and Table \ref{tab5} present performance benchmarks for representative methods in traffic parameter estimation, anomaly detection, and behavior understanding, revealing key evolutionary patterns and research trajectories.

For traffic parameter estimation (Table \ref{tab4}), methods for speed and vehicle counting have transitioned from foundational techniques like virtual sections and basic detection/tracking towards more sophisticated learning-based approaches. While accuracy has improved, a consistent challenge remains in achieving robustness across diverse environmental conditions and traffic densities without complex calibration or extensive feature engineering. Traffic anomaly detection has seen significant advancements with both weakly supervised (often MIL-based) and unsupervised (typically reconstruction or prediction-based) deep learning models. These methods demonstrate high performance on established benchmarks, yet generalizing to novel anomaly types and reducing false positives in complex, unconstrained environments are ongoing research focuses.

In behavior understanding (Table \ref{tab5}), vehicle trajectory prediction clearly shows learning-based models, particularly those leveraging recurrent, graph, and transformer architectures, outperforming traditional physics-based models in complex interactive scenarios, though physics-based approaches retain value for their interpretability in simpler contexts. For pedestrian behavior, crossing intention recognition has evolved from unimodal (trajectory or pose) to hybrid models that integrate multiple cues, consistently yielding better performance. Similarly, pedestrian trajectory prediction has progressed from social force models to deep learning techniques adept at capturing intricate social interactions. The overarching trend in behavior understanding is the increasing complexity and multimodality of models to capture nuanced interactions, though this often comes at the cost of computational load and data requirements.

These performance trends suggest future research should prioritize: (1) Enhancing model generalization and robustness to real-world variability (e.g., diverse weather, lighting, unseen behaviors) with less reliance on extensive, scenario-specific datasets; (2) Developing computationally efficient and scalable architectures suitable for real-time deployment on edge devices, particularly for complex deep learning models; (3) Improving the interpretability of black-box models and the ability to quantify prediction uncertainty, especially for safety-critical applications like anomaly detection and trajectory prediction; (4) Advancing multi-modal fusion techniques and context-aware reasoning to build more comprehensive and reliable traffic understanding systems.

Overall, while significant progress has been made, a clear direction for future research across all these tasks is the enhancement of model robustness and generalization. A critical limiting factor, particularly for traffic parameter estimation and surveillance-view traffic behavior understanding, is the scarcity of standardized, reliable, open-source, and sufficiently complex and diverse benchmark datasets. The development of such resources is crucial to facilitate fairer comparisons, rigorously evaluate generalization capabilities, and drive innovation towards solutions truly applicable in the complexities of real-world traffic environments. 

%=======================================================================
\begin{table*}[htbp]  
  \centering  
\makeatletter\setlength{\@fptop}{0pt}\makeatother  
\hspace*{-2cm}  
\begin{minipage}{\dimexpr\textwidth+4cm\relax}  
  \centering  
  \caption{Performance of current representative methods for Traffic Parameter Estimation and Traffic Anomaly Detection}  
\begin{tabular}{p{2em}|p{3em}|p{4em}|p{8em}|p{2em}|p{20em}}  
    \toprule  
    \toprule  
    \multicolumn{1}{p{2em}|}{\textbf{Task}} &   
    \multicolumn{1}{p{4.43em}|}{\textbf{Sub-task}} &   
    \multicolumn{1}{p{8em}|}{\textbf{Category (-based)}} &   
    \multicolumn{1}{p{10.5em}|}{\textbf{Method}} &   
    \multicolumn{1}{p{3.18em}|}{\textbf{Year}} &   
    \textbf{Benchmark: Metrics (\textcolor{cyan}{Source})} \\
    \midrule  
    \midrule  
    \multicolumn{1}{c|}{\multirow{19}[10]{*}{\shortstack{Traffic\\ Parameter \\Estimation}}} &   
    \multicolumn{1}{c|}{\multirow{11}[8]{*}{\shortstack{Speed \\estimation}}} &   
    \multicolumn{1}{c|}{\multirow{4}[8]{*}{\shortstack{Virtual \\ section}}} &   
    \multicolumn{1}{p{12em}|}{{Celik \& Kusetogullari \cite{celik2009solar}}} &   
    2009 &   
    Proprietary: MAE=1.23 km/h ({\small \href{https://ieeexplore.ieee.org/document/5371869}{Link}}) \\
\cmidrule{4-6}          
    &       &       &   
    \multicolumn{1}{p{8.5em}|}{{Setiyono et al. \cite{setiyono2017vehicle}}} &   
    2017 &   
    Proprietary: MAE =0.93 km/h ({\small \href{https://iopscience.iop.org/article/10.1088/1742-6596/890/1/012144}{Link}})\\
\cmidrule{4-6}          
    &       &       &   
    \multicolumn{1}{p{12em}|}{{Anandhalli et al. \cite{anandhalli2022image}}} &   
    2022 &   
    Proprietary: MAE =3.13 km/h  ({\small \href{https://link.springer.com/article/10.1007/s00138-021-01255-w}{Link}})\\
\cmidrule{4-6}          
    &       &       &   
    \multicolumn{1}{p{8.5em}|}{{Ashraf et al. \cite{ashraf2023hvd}}} &   
    2023 &   
    Proprietary: MAE =1.60 km/h  ({\small \href{https://www.sciencedirect.com/science/article/pii/S1319157823002112}{Link}})\\
\cmidrule{3-6}          
    &       &   
    \multicolumn{1}{c|}{\multirow{7}[5]{*}{\shortstack{Homography \\transformation}}} &   
    \multicolumn{1}{l|}{{{Huang \cite{huang2018traffic}}}} &  
    {2018} &   
    AI City Challenge: RMSE=3.91 (highway) ({\small \href{https://ieeexplore.ieee.org/document/8575456/}{Link}}) \\
\cmidrule{4-6}          
    &       &       &   
    \multicolumn{1}{p{8.5em}|}{{Bell et al. \cite{bell2020accurate}}} &   
    2020 &   
    Proprietary: MAE =1.53 km/h ({\small \href{https://ieeexplore.ieee.org/document/8575456/}{Link}})\\
\cmidrule{4-6}          
    &       &       &   
    \multicolumn{1}{p{8.5em}|}{{Liu et al. \cite{liu2020vision}}} &   
    2020 &   
    Proprietary: RMSE=1.85 ({\small \href{https://ieeexplore.ieee.org/document/9130874/}{Link}})\\
\cmidrule{4-6}          
    &       &       &   
    \multicolumn{1}{p{8.5em}|}{{Lashkov et al. \cite{lashkov2023edge}}} &   
    2023 &   
    BrnoCompSpeed: MAE =0.82 km/h ({\small \href{https://ieeexplore.ieee.org/document/10265296}{Link}})\\
\cmidrule{4-6}          
    &       &       &   
    \multicolumn{1}{l|}{\multirow{2}{*}{{Yohannes et al. \cite{yohannes2023improved}}}} &   
    \multirow{2}{*}{2023} &   
    BrnoCompSpeed: MSE =6.56; \\
    &       &       &       &       &   
    AI City Challenge: MSE =16.67  ({\small \href{https://ieeexplore.ieee.org/document/10015759/}{Link}})\\
\cmidrule{2-6}          
    &   
    \multicolumn{1}{c|}{\multirow{8}[5]{*}{\shortstack{Vehicle \\counting}}} &   
    \multicolumn{1}{c|}{\multirow{4}[3]{*}{\shortstack{Detection \\and tracking}}} &    
    \multicolumn{1}{p{8.5em}|}{{Song et al. \cite{song2019vision}}} &   
    2019 &   
    Freeway-vehicle: MCR = 93.2\% (cross) ({\small \href{https://etrr.springeropen.com/articles/10.1186/s12544-019-0390-4}{Link}})\\
\cmidrule{4-6}          
    &       &       &   
    \multicolumn{1}{p{8.5em}|}{{Liu et al. \cite{liu2020robust}}} &   
    2020 &   
    AI City 2020 Track-1: S1 score=93.89\% ({\small \href{https://etrr.springeropen.com/articles/10.1186/s12544-019-0390-4}{Link}})\\
\cmidrule{4-6}          
    &       &       &   
    \multicolumn{1}{p{8.5em}|}{{Majumder et al. \cite{majumder2023automated}}} &   
    2023 &   
    Proprietary: MCR = 89.59\% ({\small \href{https://www.mdpi.com/2313-433X/9/7/131}{Link}}) \\
\cmidrule{3-6}          
    &       &   
    \multicolumn{1}{c|}{\multirow{4}[3]{*}{\shortstack{Direct \\regression}}} &    
    \multicolumn{1}{p{8.5em}|}{{Zhang et al. \cite{zhang2017fcn}}} &   
    2017 &   
    TRANCOS: MAE= 4.21\%  ({\small \href{https://link.springer.com/chapter/10.1007/978-3-319-46478-7_38}{Link}})\\
\cmidrule{4-6}          
    &       &       &   
    \multicolumn{1}{p{8.5em}|}{{Yang et al. \cite{yang2021fast}}} &   
    2021 &   
    UA-DETRAC: MAE= 5.27\% ({\small \href{https://ieeexplore.ieee.org/document/9597503}{Link}})\\
\cmidrule{4-6}          
    &       &       &   
    \multicolumn{1}{p{8.5em}|}{{Guo et al. \cite{guo2023scale}}} &   
    2023 &   
    TRANCOS: MAE= 3.89\% ({\small \href{https://ieeexplore.ieee.org/document/10194472/}{Link}})\\
    \midrule  
    \multicolumn{1}{c|}{\multirow{20}[38]{*}{\shortstack{Traffic \\Anomaly\\ Detection}}} &   
    \multicolumn{1}{c|}{\multirow{11}[20]{*}{\shortstack{Weakly \\supervised}}} &   
    \multicolumn{1}{c|}{\multirow{6}[5]{*}{Classification}} &   
    \multicolumn{1}{l|}{{{Deep-Cascade \cite{sabokrou2017deep}}}} &   
    {2017} &   
    UCSDped1/ped2: EER=9.1\% \& 8.2\% ({\small \href{https://ieeexplore.ieee.org/document/7858798}{Link}}) \\
\cmidrule{4-6}          
    &       &       &   
    \multicolumn{1}{p{8.5em}|}{{ConvLSTM \cite{lu2020new}}} &   
    2019 &   
    Proprietary: ACC=87.78\% ({\small \href{https://onlinelibrary.wiley.com/doi/10.1155/2020/8848874}{Link}})\\
\cmidrule{4-6}          
    &       &       &   
    \multicolumn{1}{p{8.5em}|}{{GCN-AD \cite{zhong2019graph}}} &   
    2020 &   
    Shanghai Tech: AUC=84.44\% ({\small \href{https://arxiv.org/abs/1903.07256}{Link}})\\
\cmidrule{4-6}          
    &       &       &   
    \multicolumn{1}{p{8.5em}|}{{MIST \cite{feng2021mist}}} &   
    2021 &   
    Shanghai Tech: AUC=94.83\% ({\small \href{https://arxiv.org/abs/2104.01633}{Link}})\\
\cmidrule{4-6}          
    &       &       &   
    \multicolumn{1}{p{8.5em}|}{{Two stream \cite{zhou2023appearance}}} &   
    2023 &   
    CDD dataset: AUC=0.96 ({\small \href{https://ieeexplore.ieee.org/document/10197171/}{Link}})\\
\cmidrule{3-6}          
    &       &   
    \multicolumn{1}{c|}{\multirow{5}[10]{*}{Scoring}} &   
    \multicolumn{1}{p{8.5em}|}{{C3D+ MIL \cite{sultani2018real}}} &   
    2018 &   
    Proprietary: AUC=75.41\% ({\small \href{https://arxiv.org/abs/1801.04264}{Link}})\\
\cmidrule{4-6}          
    &       &       &   
    \multicolumn{1}{p{8.5em}|}{{TAN+ MIL \cite{zhu2019motion}}} &   
    2019 &   
    UCF Crime: AUC= 79.0\% ({\small \href{https://arxiv.org/abs/1907.10211}{Link}})\\
\cmidrule{4-6}          
    &       &       &   
    \multicolumn{1}{p{12em}|}{{\shortstack[l]{Self-reasoning\\ framework \cite{zaheer2020self}}}} &   
    {2020} &   
    \shortstack[l]{UCF-Crime: AUC=79.54\%; \\ Shanghai Tech: AUC= 84.16\%} ({\small \href{https://ieeexplore.ieee.org/document/9204830/}{Link}})\\
\cmidrule{4-6}          
    &       &       &   
    \multicolumn{1}{p{8.5em}|}{{NTCN-ML \cite{shao2023video} }} &   
    2023 &   
    \shortstack[l]{UCF-Crime: AUC= 85.1\%; \\Shanghai Tech: AUC= 95.3\%} ({\small \href{https://ieeexplore.ieee.org/document/9204830/}{Link}})\\
\cmidrule{4-6}          
    &       &       &   
    \multicolumn{1}{p{8.5em}|}{{MC-MIL \cite{pereira2024mc} }} &   
    2023 &   
    PETS 2009: AUC= 95.39\% ({\small \href{https://link.springer.com/article/10.1007/s00521-024-09611-3}{Link}})\\
\cmidrule{2-6}          
    &   
    \multicolumn{1}{c|}{\multirow{9}[18]{*}{Unsupervised}} &   
    \multicolumn{1}{c|}{\multirow{5}[10]{*}{Reconstruction}} &   
    \multicolumn{1}{p{8.5em}|}{{Conv-AE \cite{hasan2016learning}}} &   
    2016 &   
    \shortstack[l]{UCSDped1/UCSDped2: AUC=92.7\%/90.8\%; \\CUHK Avenue: AUC=70.2\% ({\small \href{https://arxiv.org/abs/1604.04574}{Link}})} \\
\cmidrule{4-6}          
    &       &       &   
    \multicolumn{1}{p{8.5em}|}{{MemAE \cite{gong2019memorizing}}} &   
    2019 &   
    \shortstack[l]{UCSDped2: AUC: 94.1\%;  \\Shanghai Tech: AUC= 71.2\%} ({\small \href{https://arxiv.org/abs/1904.02639}{Link}})\\
\cmidrule{4-6}          
    &       &       &   
    \multicolumn{1}{p{8.5em}|}{{STAE \cite{deepak2021residual}}} &   
    2021 &   
    \shortstack[l]{UCSDped2: AUC=83\%; \\ CUHK Avenue: AUC=82\%} ({\small \href{https://link.springer.com/article/10.1007/s11760-020-01740-1}{Link}})\\
\cmidrule{4-6}          
    &       &       &   
    \multicolumn{1}{p{12em}|}{Hybrid CNN-VAE \cite{santhosh2020anomaly}} &   
{2021} &   
    T15: ACC=99.0\%; QMUL: ACC=97.3\% ({\small \href{https://ieeexplore.ieee.org/document/9531567}{Link}})\\
\cmidrule{4-6}          
    &       &       &   
    \multicolumn{1}{p{8.5em}|}{{\shortstack[l]{LSTM autoencoder + \\ AL + ATD \cite{zhou2022vision}}}} &   
    2022 &   
    Proprietary: ACC=97.0\% ({\small \href{https://link.springer.com/article/10.1007/s00521-022-07335-w}{Link}})\\
\cmidrule{3-6}          
    &       &   
    \multicolumn{1}{c|}{\multirow{4}[8]{*}{Prediction}} &   
    \multicolumn{1}{p{8.5em}|}{\multirow{2}[2]{*}{HF2-VAD \cite{liu2021hybrid}}} &   
    \multirow{2}[2]{*}{2020} &   
    Shanghai Tech: AUC= 76.2\%; UCSDped2: AUC= 99.3\%; CUHK Avenue: AUC= 91.1\% ({\small \href{https://arxiv.org/abs/2108.06852}{Link}})\\
\cmidrule{4-6}          
    &       &       &   
    \multicolumn{1}{p{8.5em}|}{{ROADMAP \cite{wang2021robust}}} &   
    2022 &   
    \shortstack[l]{Shanghai Tech: AUC=76.6\%; \\ CUHK Avenue: AUC= 88.3\%} ({\small \href{https://arxiv.org/abs/2011.02763}{Link}})\\
\cmidrule{4-6}          
    &       &       &   
    \multicolumn{1}{p{12em}|}{{ST-Transformer \cite{tran2024transformer} }} &   
    2024 &   
    \shortstack[l]{UIT-ADrone: AUC= 65.45\%; \\ Drone-Anomaly: AUC=67.80\%} ({\small \href{https://ieeexplore.ieee.org/document/10466765/}{Link}})\\
    \bottomrule  
\end{tabular}%  
  \label{tab4}%  
  \end{minipage}  
\end{table*}%

\begin{table*}[htbp]  
\centering  
\makeatletter\setlength{\@fptop}{0pt}\makeatother  
\hspace*{-2cm}  
\begin{minipage}{\dimexpr\textwidth+3cm\relax}  
\centering  
\caption{Performance of current representative methods for Behavior Understanding}  
\begin{tabular}{p{2em}|p{3em}|p{5em}|p{8em}|p{2em}|p{20em}}  
    \toprule  
    \toprule  
    \multicolumn{1}{p{2em}|}{\textbf{Task}} &   
    \multicolumn{1}{p{4.43em}|}{\textbf{Sub-task}} &   
    \multicolumn{1}{p{8em}|}{\textbf{Category (-based)}} &   
    \multicolumn{1}{p{12.5em}|}{\textbf{Method}} &   
    \multicolumn{1}{p{3.18em}|}{\textbf{Year}} &   
    \textbf{Benchmark: Metrics  (\textcolor{cyan}{Source})} \\
    \midrule  
    \midrule  
    \multicolumn{1}{c|}{\multirow{40}[20]{*}{\shortstack{Behavior\\ Understanding}}} &   
    \multicolumn{1}{c|}{\multirow{6}[10]{*}{\shortstack{Vehicle\\ trajectory\\ prediction}}} &   
    \multicolumn{1}{c|}{\multirow{2}[2]{*}{Physics}} &   
    \multicolumn{1}{p{8.5em}|}{{IMMTP \cite{xie2017vehicle}}} &   
    2017 &   
    Proprietary: APE=1.55m (PT =8s) ({\small \href{https://ieeexplore.ieee.org/document/8186191/}{Link}})\\
\cmidrule{4-6}          
    &       &       &   
    \multicolumn{1}{p{8.5em}|}{{Anderson et al. \cite{anderson2021kinematic}}} &   
    2021 &   
    \shortstack[l]{NGSIM: ADE=3.14m, RMSE=4.08\%; \\ highD: ADE=1.51m, RMSE=1.92\%} ({\small \href{https://arxiv.org/abs/2103.16673}{Link}})\\
\cmidrule{3-6}          
    &       &   
    \multicolumn{1}{c|}{\multirow{4}[8]{*}{Learning}} &   
    \multicolumn{1}{p{8.5em}|}{{DeepTrack \cite{katariya2022deeptrack}}} &   
    2022 &   
    NGSIM: ADE=2.01m, FDE=3.25m ({\small \href{https://arxiv.org/abs/2108.00505}{Link}})\\
\cmidrule{4-6}          
    &       &       &   
    \multicolumn{1}{p{8.5em}|}{{D2-TPred \cite{zhang2022d2}}} &   
    2022 &   
    VTP-TL: ADE=16.9 pixel, FDE=34.6 pixel ({\small \href{hhttps://arxiv.org/abs/2207.10398}{Link}})\\
\cmidrule{4-6}          
    &       &       &   
    \multicolumn{1}{p{8.5em}|}{{DACR-AMTP \cite{cong2023dacr}}} &   
    2023 &   
    \shortstack[l]{NGSIM: ADE=1.61m, FDE=3.31m; \\ highD: ADE=0.76m, FDE=1.69m} ({\small \href{https://ieeexplore.ieee.org/document/10269660/}{Link}})\\
\cmidrule{4-6}          
    &       &       &   
    \multicolumn{1}{p{8.5em}|}{{VT-Former \cite{pazho2024vt}}} &   
    2024 &   
    \shortstack[l]{NGSIM: ADE= 2.10m, FDE=4.91m; \\ CHD dataset: ADE=25.33 pixel, FDE=88.99 pixel} ({\small \href{https://arxiv.org/abs/2311.06623}{Link}})\\
\cmidrule{2-6}          
    &   
    \multicolumn{1}{c|}{\multirow{14}[20]{*}{\shortstack{Pedestrian\\ crossing\\ intention\\ recognition}}} &   
    \multicolumn{1}{c|}{\multirow{2}[4]{*}{Trajectory}} &   
    \multicolumn{1}{p{12em}|}{\multirow{2}[2]{*}{Goldhammer et al. \cite{goldhammer2019intentions}}} &   
    \multirow{2}[2]{*}{2019} &   
    Proprietary: ACC=98.6\% (Waiting), 77.1\% (Starting), 88.1\%(Walking), Stopping (60.9\%) ({\small \href{https://ieeexplore.ieee.org/document/8748209}{Link}})\\
\cmidrule{4-6}          
    &       &       &   
    \multicolumn{1}{p{8.5em}|}{{PIE$_{int}$ \cite{rasouli2017they}}} &   
    2019 &   
    PIE: ACC=69\%, F1-score=79\% ({\small \href{https://ieeexplore.ieee.org/document/9008118/}{Link}})\\
\cmidrule{3-6}          
    &       &   
    \multicolumn{1}{c|}{\multirow{3}[6]{*}{Pose}} &   
    \multicolumn{1}{p{8.5em}|}{{Fang et al. \cite{fang2019intention}}} &   
    2020 &   
    JAAD: ACC=88\% ({\small \href{https://arxiv.org/abs/1910.03858}{Link}})\\
\cmidrule{4-6}          
    &       &       &   
    \multicolumn{1}{p{8.5em}|}{{Xu et al. \cite{xu2021action}}} &   
    2022 &   
    3D-HPT: ACC=88.34\% (Cross-subject), ACC=89.62\% (Cross-view) ({\small \href{https://ieeexplore.ieee.org/document/9660768/}{Link}})\\
\cmidrule{4-6}          
    &       &       &   
    \multicolumn{1}{p{8.5em}|}{{Zhang et al. \cite{zhang2021pedestrian}}} &   
    2022 &   
    Proprietary: AUC=84.1\% (2 sec) ({\small \href{https://ieeexplore.ieee.org/document/9423518}{Link}})\\
\cmidrule{3-6}          
    &       &   
    \multicolumn{1}{c|}{\multirow{9}[5]{*}{Hybrid}} &   
    \multicolumn{1}{l|}{\multirow{2}[2]{*}{{TrouSPI-Net \cite{gesnouin2021trouspi}}}} &   
    \multirow{2}[2]{*}{2021} &   
    PIE: ACC=88\%, AUC=88\%, F1-score=80\%; \\
    &       &       &       &       &   
    JAAD: ACC=85\%, AUC=73\%, F1-score=56\% ({\small \href{https://arxiv.org/abs/2109.00953}{Link}})\\
\cmidrule{4-6}          
    &       &       &   
    \multicolumn{1}{l|}{\multirow{2}[2]{*}{{PCPA \cite{kotseruba2021benchmark}}}} &   
    \multirow{2}[2]{*}{2021} &   
    PIE: ACC=87\%, AUC=86\%, F1-score=77\%; \\
    &       &       &       &       &   
    JAAD: ACC=85\%, AUC=86\%, F1-score=68\% ({\small \href{https://ieeexplore.ieee.org/document/9423436}{Link}})\\

\cmidrule{4-6}          
    &       &       &   
    \multicolumn{1}{p{8.5em}|}{{PIP-Net \cite{azarmi2024pip}}} &   
    2024 &   
    PIE: ACC=91\%, AUC=90\%, F1-score=84\% ({\small \href{https://arxiv.org/abs/2402.12810}{Link}})\\
\cmidrule{4-6}          
    &       &       &   
    \multicolumn{1}{l|}{\multirow{2}[2]{*}{{PedCMT \cite{chen2024pedestrian}}}} &   
    \multirow{2}[2]{*}{2024} &   
    PIE: ACC=93\%, AUC=92\%, F1-score=87\%; \\
    &       &       &       &       &   
    JAAD: ACC=88\%, AUC=77\%, F1-score=65\% ({\small \href{https://ieeexplore.ieee.org/document/10507743/}{Link}})\\
\cmidrule{2-6}          
    &   
    \multicolumn{1}{c|}{\multirow{13}[13]{*}{\shortstack{Pedestrian\\ trajectory\\ prediction}}} &   
    \multicolumn{1}{c|}{Physics} &   
    \multicolumn{1}{p{12em}|}{{W/CDM-MSFM \cite{zhang2021pedestrian}}} &   
    2021 &   
    Proprietary: FDE=0.136 m ({\small \href{https://ieeexplore.ieee.org/document/9036079/}{Link}})\\
\cmidrule{3-6}          
    &       &   
    \multicolumn{1}{c|}{\multirow{12}[11]{*}{Learning}} &   
    \multicolumn{1}{l|}{\multirow{2}[2]{*}{{Social LSTM \cite{alahi2014socially}}}} &   
    \multirow{2}[2]{*}{2016} &   
    ETH: ADE=1.09m, FDE=2.35m; \\
    &       &       &       &       &   
    HOTEL: ADE=0.79m, FDE=1.76m ({\small \href{https://arxiv.org/abs/1803.10892/}{Link}})\\
\cmidrule{4-6}          
    &       &       &   
    \multicolumn{1}{l|}{\multirow{2}[2]{*}{{Social GAN \cite{gupta2018social}}}} &   
    \multirow{2}[2]{*}{2018} &   
    ETH: ADE=0.60m, FDE=1.19m; \\
    &       &       &       &       &   
    HOTEL: ADE=0.67m, FDE=1.37m ({\small \href{https://arxiv.org/abs/1803.10892/}{Link}})\\
\cmidrule{4-6}          
    &       &       &   
    \multicolumn{1}{l|}{\multirow{2}[2]{*}{{Social STGCN \cite{mohamed2020social}}}} &   
    \multirow{2}[2]{*}{2020} &   
    ETH: ADE=0.64m, FDE=1.11m; \\
    &       &       &       &       &   
    HOTEL: ADE=0.49m, FDE=0.85m ({\small \href{https://arxiv.org/abs/2002.11927}{Link}})\\
\cmidrule{4-6}          
    &       &       &   
    \multicolumn{1}{l|}{\multirow{2}[2]{*}{{SGCN \cite{shi2021sgcn}}}} &   
    \multirow{2}[2]{*}{2021} &   
    ETH: ADE=0.63m, FDE=1.03m; \\
    &       &       &       &       &   
    HOTEL: ADE= 0.32m, FDE=0.55m ({\small \href{https://arxiv.org/abs/2104.01528}{Link}})\\
\cmidrule{4-6}          
    &       &       &   
    \multicolumn{1}{l|}{\multirow{2}[2]{*}{{SSAGCN \cite{lv2023ssagcn}}}} &   
    \multirow{2}[2]{*}{2023} &   
    ETH: ADE=0.3m, FDE=0.59m; \\
    &       &       &       &       &   
    HOTEL: ADE=0.22m, FDE=0.42m  ({\small \href{https://ieeexplore.ieee.org/document/10063206}{Link}})\\
\cmidrule{4-6}          
    &       &       &   
    \multicolumn{1}{l|}{\multirow{2}[1]{*}{{TUTR \cite{shi2023trajectory}}}} &   
    \multirow{2}[1]{*}{2023} &   
    ETH: ADE=0.40m, FDE=0.61m; \\
    &       &       &       &       &   
    HOTEL: ADE=0.11m, FDE=0.18m ({\small \href{https://ieeexplore.ieee.org/document/10378344/}{Link}})\\
    \bottomrule  
\end{tabular}%  

  \label{tab5}% 
      \vspace{1em} % 在表格和注释之间添加一些空间  
    \begin{justify} % 确保文本两端对齐  
    \small % 使注释字体略小  
    \textbf{Note:} APE =  Average Prediction Error, PT = Prediction Time, RMSE = Root Mean Square Error, ADE = Average Displacement Error, FDE = Final Displacement Error, AUC = Area Under the Curve, ACC = Accuracy  
    \end{justify}  
  \end{minipage}  
\end{table*}%

\section{Limitation Analysis and Future Outlook}

\subsection{Limitation Overview}

\begin{figure}[!tb]
  \centering
  \includegraphics[width=\textwidth]{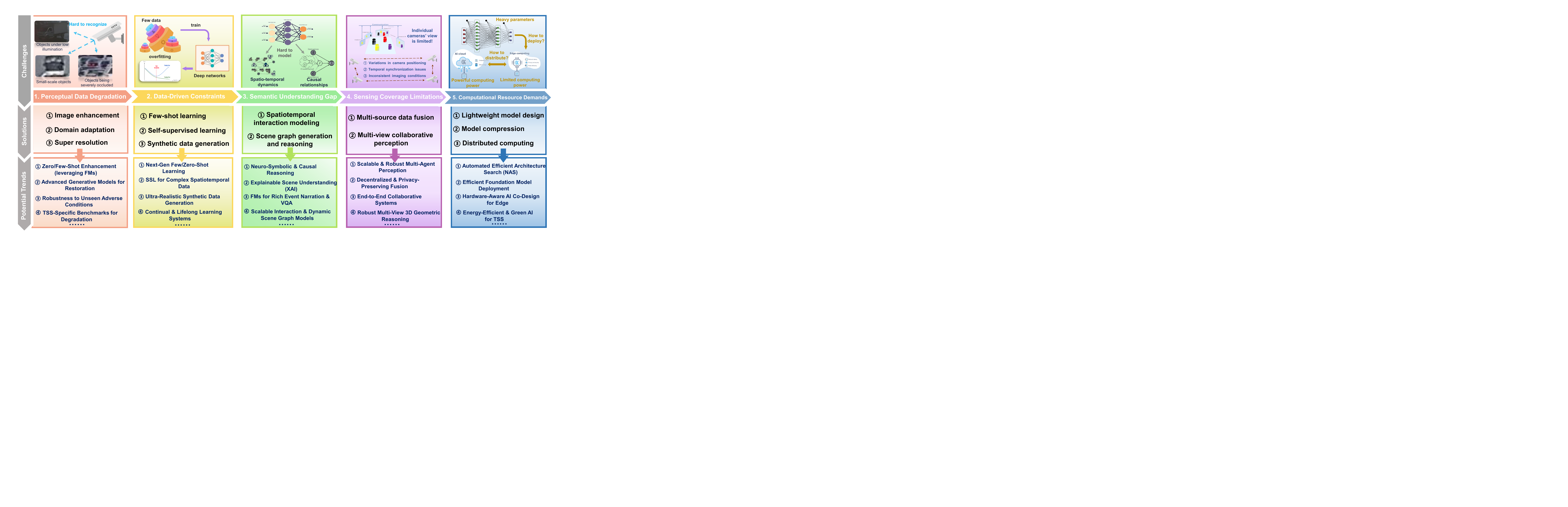}
  \caption{ Overview of challenges, solutions, and potential trends for vision technologies in traffic surveillance}
  \label{fig14}
\end{figure}

Building upon the comprehensive analysis of strengths and limitations of existing methods across different tasks, several fundamental limitations persist in TSS despite continued advances in vision technologies (as shown in Figure \ref{fig14}):

a)	\textbf{Perceptual data degradation}:  The foundational quality and completeness of visual data are frequently compromised in complex traffic scenarios. Factors like high traffic density, congestion, nighttime conditions, and adverse weather lead to degraded or incomplete sensory input. Consequently, existing perception algorithms, particularly those reliant on consistent feature extraction or clear visual cues \cite{zhou2023monitoring,zhou2023appearance,zhou2023all} , often falter, resulting in frequent false positives/negatives and a significant reduction in overall system reliability and safety.

b)	\textbf{Data-driven learning constraints}: The predominant deep neural network paradigm, while powerful, is inherently constrained by significant data-related challenges. The necessity for large-scale, meticulously annotated datasets is particularly problematic in traffic surveillance due to privacy concerns and the sheer effort involved. Moreover, the natural rarity of critical traffic events (e.g., accidents, violations) creates severe data imbalance. As a result, current supervised learning approaches \cite{zhou2022automated, kamenou2023meta} are often hindered in their ability to generalize robustly, learn rapidly from limited samples, or adapt effectively to novel environments without extensive retraining.

c)	\textbf{Semantic understanding gap}: While proficient at feature-based detection and recognition, existing deep learning models \cite{zou2023object,luo2021multiple} typically operate with a limited grasp of the broader traffic context. They largely lack the capability for commonsense reasoning, struggling to interpret the intricate relationships between objects, their interactions with the environment, and the underlying causal dynamics within complex traffic scenarios. This gap prevents systems from moving beyond simple perception to a more holistic and predictive understanding of traffic behavior.

d)	\textbf{Sensing coverage limitations}:  The inherent field-of-view restrictions of individual cameras significantly limit their capacity for effective monitoring of large-scale or complex traffic environments. While multi-camera systems can theoretically extend coverage, their practical effectiveness is often undermined by substantial challenges. These include cross-camera object association, spatio-temporal alignment, and robust data fusion, especially given variations in camera positioning, synchronization issues, and inconsistent imaging conditions. These issues directly impede the creation of a cohesive and comprehensive situational awareness picture.

e) \textbf{Computational resource demands}: Many contemporary deep learning models achieving state-of-the-art accuracy in traffic surveillance demand substantial computational resources. This is particularly true for complex architectures required for robust perception and understanding. The imperative for real-time processing in dynamic traffic environments often clashes with the computational intensity of these models, posing a significant barrier to their deployment, especially on resource-constrained edge devices. This burden translates directly into increased energy consumption, higher hardware costs, and ultimately limits the scalability and widespread practical adoption of the most advanced TSS solutions.

\subsection{Current Solutions and Potential Trends}

To address the limitations outlined in Section 5.1, researchers have proposed and developed a spectrum of technical solutions and methodological innovations, as illustrated in Figure \ref{fig14}. While some of these areas represent established research fields with significant advancements, their application and adaptation to the unique complexities of TSS continue to present challenges and offer avenues for future innovation. This section will review key approaches, critically assess their current standing and limitations within TSS, and provide a forward-looking perspective on emerging trends and necessary advancements.

\subsubsection{Advanced perception enhancement}
Advanced perception enhancement techniques, including \textit{image enhancement}, \textit{domain adaptation}, and \textit{super-resolution}, are vital for improving visual data quality under challenging TSS conditions.

\textit{Image enhancement} seeks to improve degraded image quality, with modern approaches utilizing Generative Adversarial Networks (GANs) (e.g., EnlightenGAN \cite{jiang2021enlightengan}, N2DGAN \cite{schutera2020night} for low-light; CoMoGAN \cite{pizzati2021comogan}, IA-GAN \cite{zhou2023all} for day-to-night translation) and diffusion models (e.g., LightDiff \cite{li2024light}), alongside methods addressing specific conditions like domain shifts in lighting \cite{li2021domain} or adverse weather \cite{xiao2022image, chen2023learning}. However, a critical challenge in TSS is achieving robust generalization across diverse, unconstrained adverse conditions without significant computational overhead or artifacts, especially as powerful diffusion models often face deployment feasibility issues due to high costs \cite{zhou2024synthesizing}. Future research should therefore focus on lightweight yet robust models, unsupervised/self-supervised enhancement requiring less paired data, creating TSS-specific benchmarks for diverse adverse conditions, and integrating foundation models for zero/few-shot enhancement in novel scenarios.

\textit{Domain adaptation (DA)} aims to mitigate performance drops when models encounter new domains by adapting feature representations \cite{chen2018domain, chen2020harmonizing, munir2023domain}, employing techniques like dual-level adaptation \cite{chen2018domain}, multi-level calibration \cite{chen2020harmonizing}, and uncertainty-guided methods \cite{munir2023domain}. Despite these advances, seamless adaptation between significantly different TSS domains (e.g., clear day vs. snowy night) remains difficult, particularly for fine-grained tasks, and the “black-box” nature of some DA methods can hinder interpretability. Consequently, future work should prioritize more explainable DA, continuous DA for evolving traffic environments, multi-target DA, and the development of TSS-specific DA benchmarks to systematically evaluate robustness against various domain shifts.

\textit{Super-resolution (SR)} reconstructs high-resolution images from low-resolution inputs, evolving from basic enhancement \cite{zhou2023monitoring} to advanced structure restoration \cite{zamir2022restormer}, with innovations including self-supervised \cite{pan2020self} and GAN-based SR \cite{basak2024vehicle}. However, TSS faces unique challenges: ensuring SR improvements translate to better downstream performance (e.g., vehicle detection, license plate recognition) rather than mere perceptual enhancement; computational constraints for real-time processing in edge devices; and the “perception-performance gap” where visually appealing results may not improve analytical accuracy for distant or occluded objects. Future research should prioritize developing SR methods optimized for specific TSS tasks (e.g., detection, vehicle Re-ID), lightweight architectures suitable for edge deployment, evaluation frameworks that measure impact on end-to-end system performance, and adaptive SR that adjusts enhancement strategies based on traffic scene complexity and object importance.

\subsubsection{Efficient learning paradigms}
Efficient learning paradigms, primarily \textit{few-shot learning (FSL)}, \textit{self-supervised learning (SSL)}, and \textit{synthetic data generation}, are crucial for mitigating the heavy data dependency of deep learning in TSS. While established fields, their robust application in diverse TSS contexts is an active research area.

\textit{Few-shot learning (FSL)} enables models to learn from minimal examples, evolving from metric learning \cite{vinyals2016matching} to meta-learning frameworks \cite{zhang2022meta}, with TSS applications in equipment detection \cite{zhou2022automated} and vehicle re-identification \cite{kamenou2023meta}. However, FSL performance can be sensitive to “shot” selection and may struggle with generalization to novel or fine-grained TSS categories; curating representative support sets for deployment also presents a hidden cost. Future directions therefore include developing more robust FSL methods, exploring meta-learning for continual adaptation, combining FSL with active learning, and creating standardized FSL benchmarks for diverse TSS tasks.

\textit{Self-supervised learning (SSL)} extracts features from unlabeled data via pretext tasks, progressing from rotation prediction \cite{gidaris2018unsupervised} to contrastive learning and masked image modeling \cite{he2022masked}, with TSS applications in anomaly detection \cite{huang2021abnormal} and multi-task SSL \cite{barbalau2023ssmtl++}. A key challenge is designing pretext tasks that effectively capture complex spatio-temporal traffic semantics, as transferability of SSL-learned representations to diverse downstream TSS tasks and geolocations is not guaranteed, with real-world applicability often hinging on pretext-target task alignment. Future advancements should explore SSL pretext tasks for traffic dynamics, robust SSL across varying conditions, combining SSL with weak supervision or domain knowledge, and leveraging pre-trained foundation models as powerful starting points for SSL in TSS.

\textit{Synthetic data generation} creates large, auto-labeled datasets using simulation \cite{luo2023simulation}, domain randomization \cite{yue2019domain}, physics-based rendering \cite{li2023novel}, and generative models \cite{richter2022enhancing}, demonstrated in accident video generation \cite{vijay2022detection} and realism enhancement \cite{richter2022enhancing}. The primary limitation remains the “domain gap” between synthetic and real data, which can degrade real-world performance, while ensuring diversity and realism for complex interactions and rare events is difficult \cite{zhou2024synthesizing}; over-reliance on synthetic data can lead to poor deployment outcomes. Future efforts should focus on reducing this gap (e.g., advanced domain randomization, GAN-based refinement), developing realistic traffic simulation platforms, methodologies for optimally blending synthetic/real data, and creating benchmark datasets that enable systematic evaluation of synthetic-to-real transfer performance.

\subsubsection{Knowledge-Enhanced Understanding}
Knowledge-enhanced approaches, focusing on \textit{spatiotemporal interaction modeling} and \textit{scene graph generation and reasoning}, aim for genuine comprehension of the evolving interactions and complex behaviors of vehicles and pedestrians within the traffic environment, moving beyond mere pattern recognition.

\textit{Spatiotemporal interaction modeling} captures dynamic relationships between traffic participants, often using GNNs \cite{ling2024pedast}, Attention Mechanisms \cite{azarmi2024pip}, or Transformers \cite{chen2024pedestrian} for tasks like pedestrian intent and trajectory prediction. Modeling complex, multi-agent interactions in dense, unpredictable traffic with high performance and real-time constraints remains difficult, and ensuring models capture causality, not just correlation, is a hurdle for real-world applicability in safety-critical systems. Future directions should include developing more scalable and interpretable interaction models, incorporating explicit reasoning modules (e.g., traffic rules, human behavior models), combining these with probabilistic approaches for uncertainty, and creating datasets with rich annotations of multi-agent interactions and outcomes.

\textit{Scene graph generation and reasoning} constructs  semantic representations of scenes, which, while applied in VQA \cite{koner2021graphhopper}, multimedia \cite{curry2022multimodal}, and captioning \cite{yang2020auto}, are less mature in TSS. Key challenges include capturing temporal dynamics and evolving relationships in traffic scenarios, multi-granularity reasoning across spatial-temporal scales, handling uncertainty in traffic behaviors, and efficient reasoning over large, dynamic graphs, with deployment feasibility for real-time applications being a major concern. Future research should explore temporal scene graphs with memory mechanisms, integration with foundation models for zero-shot traffic reasoning, probabilistic graphs incorporating uncertainty quantification, and developing TSS benchmarks for comprehensive scene graph understanding.

\subsubsection{Cooperative sensing frameworks}
Addressing limited sensing coverage involves \textit{multi-source data fusion} and \textit{multi-view collaborative perception} \cite{xu2018pointfusion}, though widespread, robust, and standardized implementation of collaborative perception in TSS is still evolving.

\textit{Multi-source data fusion} combines various data types like video/images \cite{li2024scene}, text \cite{ali2021traffic}, social media \cite{li2024scene}, mobile signaling \cite{liu2023telecomtm}, street/satellite imagery \cite{swerdlow2024street, wang2024deep}, and structured data \cite{song2019mapping}, using diverse methods \cite{zhu2018urban, wang2021traffic, peng2020spatial} for applications like traffic state estimation \cite{mishra2024integrating}. Challenges include handling data heterogeneity across modalities, ensuring temporal synchronization and spatial calibration, managing inconsistent data quality and sensor reliability in dynamic traffic environments, real-time processing constraints for timely decision-making, and establishing rigorous evaluation frameworks for fusion performance across diverse TSS scenarios. Future trends point towards adaptive fusion algorithms that handle missing or degraded sensor inputs, intelligent sensor selection and weighting based on real-time quality assessment, standardized multi-modal TSS data frameworks with common interfaces, and comprehensive datasets with temporally aligned, diverse modalities under varying environmental conditions.

\textit{Multi-view collaborative perception} leverages multiple, primarily camera-based, viewpoints to create a superior 3D scene representation, overcoming single-view occlusions and field-of-view limits through data, feature, or result-level fusion \cite{arnold2020cooperative, chen2024end, su2023uncertainty, xu2022opv2v, yu2022dair, gao2024survey}. Key perceptual challenges involve robustly aligning features across varied, dynamic perspectives—often with imperfect calibration—requiring advanced geometric and view-invariant learning. Effective communication (sufficient bandwidth, low latency) is crucial for sharing rich visual data, but the ultimate aim is this enhanced collective visual understanding. Future perception-centric work should target: (1) robust cross-view feature matching and 3D reconstruction resilient to calibration and viewpoint shifts; (2) intelligent view/feature selection for task-optimized fusion; and (3) sophisticated multi-view datasets and simulations that rigorously test collaborative visual algorithms against real-world geometric complexities like inter-view occlusions and varied sensor arrangements, fostering a shared visual reality.

\subsubsection{Efficient computing frameworks}
Efficient computing through \textit{lightweight model design}, \textit{model compression}, and \textit{distributed computing strategies} is paramount for real-world applicability of advanced AI in TSS, especially on edge devices.

\textit{Lightweight model design} creates efficient architectures using techniques like depth-wise separable convolutions \cite{mazhar2023rethinking}, channel attention \cite{zheng2023lightweight}, and NAS \cite{liu2021survey}, exemplified by MobileViT \cite{mehta2021mobilevit}, EfficientFormer \cite{li2022efficientformer}, Deeptrack \cite{katariya2022deeptrack}, and LightMOT \cite{guo2023lightmot}. The main challenge is achieving optimal accuracy-efficiency trade-offs for complex TSS tasks, as lightweight models often struggle with challenging scenarios like occlusion, low resolution, and adverse weather conditions. Future directions include TSS-specific NAS frameworks, adaptive models that dynamically adjust complexity based on scene complexity and resource availability, and efficient vision Transformer architectures optimized for traffic video analysis.

\textit{Model compression} reduces model size via quantization \cite{yang2019cdeeparch}, pruning \cite{molchanov2019importance}, and knowledge distillation \cite{mirzadeh2020improved}, with innovations like hardware-aware \cite{xiao2023haloc} and dynamic compression \cite{hu2023dynamic}. However, aggressive compression can significantly degrade performance on critical TSS tasks, compression strategies show high variability across different model architectures and TSS applications, and standardized evaluation protocols for compressed models in TSS are lacking. Future work should explore automated compression pipelines tailored for TSS workloads, compression methods that maintain robustness to domain shifts and adverse conditions, and co-design approaches that integrate compression considerations into initial architecture development.

\textit{Distributed computing strategies} optimize resource use via edge-cloud collaboration \cite{wang2024end}, distributed intelligent systems \cite{chen2019cooper}, adaptive computation offloading \cite{mustafa2024deep}, and federated learning \cite{qi2024model}. Key challenges include managing communication overhead in bandwidth-constrained environments, ensuring data consistency and security across distributed nodes, designing effective task allocation strategies for heterogeneous edge devices, and balancing latency requirements with computational efficiency. Future trends involve enhanced edge intelligence with specialized accelerators, privacy-preserving federated learning frameworks for diverse TSS deployments, adaptive resource management systems, and efficient distributed deployment strategies for large foundation models through model partitioning and edge-specific optimization.

\subsection{Foundation Model Prospects}

Foundation models (FMs), also known as large models, have recently transformed the landscape of artificial intelligence. These include Large Language Models (LLMs, e.g., ChatGPT 3.5), Large Vision Models (LVMs, e.g., SAM \cite{kirillov2023segment}), and Vision-Language Models (VLMs, e.g., CLIP \cite{radford2021learning}, GPT-4V) that combine both capabilities, all demonstrating unprecedented capabilities in their respective domains. These models, pre-trained on massive datasets, exhibit remarkable zero-shot learning abilities, strong generalization, and sophisticated reasoning capabilities across diverse tasks. 

In the context of TSS, the emergence of FMs presents unique opportunities due to their distinctive advantages: the ability to understand complex visual scenes, reason about spatial-temporal relationships, and transfer knowledge across different traffic scenarios. These capabilities directly address several fundamental challenges in current TSS, particularly in alleviating data-driven learning constraints and bridging the semantic understanding gap. Additionally, foundation world models (FWMs) such as SORA\footnote{https://openai.com/sora/}, which can learn and simulate the dynamics of traffic environments, offer promising potential for controlled data and scene generation in TSS for enhancing visual perception capabilities, particularly in rare event detection and complex scenario understanding.

Therefore, the subsequent sections will elaborate on four key aspects: (1) towards data-efficient learning, (2) bridging semantic gaps, (3) scene generation via FWMs, and (4) challenges and solutions.

\subsubsection{Towards data-efficient learning}

FMs demonstrate remarkable capabilities in mitigating data dependency through their pre-trained knowledge and transfer learning abilities. Their few-shot and zero-shot learning capabilities are particularly valuable for TSS applications where labeled data is scarce or difficult to obtain. For instance, in traffic object detection, models like SAM \cite{kirillov2023segment} and CLIP \cite{radford2021learning} have shown the ability to segment and detect various traffic participants with minimal fine-tuning, reducing the annotation burden for specific deployment scenarios \cite{shokri2024proposing,zhou2024teaching} and enhancing the transferability and flexibility of detectors \cite{zhao2024tsclip}. In traffic anomaly detection, where abnormal events are naturally rare, FMs can leverage their pre-trained knowledge to identify unusual patterns even with limited examples \cite{lohner2024enhancing}. Moreover, their transfer learning capabilities enable rapid adaptation to new traffic environments \cite{zhao2024tsclip} or object categories \cite{wu2024towards}, addressing the challenge of dataset bias and environmental variations. For instance, open-vocabulary classification and detection capabilities in TSS applications enable models to identify novel traffic participants not present in the training set, such as emerging mobility devices, region-specific vehicles (like tuk-tuks in Southeast Asia), and temporary traffic facilities.

\subsubsection{Bridging semantic gaps}

FMs excel at understanding complex semantic relationships and contextual information, offering unprecedented opportunities for high-level traffic scene understanding. Their sophisticated reasoning capabilities, typically implemented through Visual Question Answering (VQA) or dense captioning mechanisms \cite{tom2023reading,abu2024using}, enable better interpretation of spatial-temporal relationships and complex interactions among traffic participants. This VQA-based approach has proven particularly effective in safety-critical events (SCEs) understanding, where models can analyze and describe complex scenarios such as crashes, near-crashes, and traffic violations. Additionally, some recent studies \cite{guo2024cfmmc,zhang2023study} have explored FMs' capabilities in performing higher-order tasks such as accident cause analysis and counterfactual reasoning, where models can infer potential causes of accidents, generate alternative scenarios (“what-if” analysis), and propose preventive measures based on comprehensive scene understanding and causal reasoning capabilities.

Moreover, the multi-modal processing capabilities of FMs enable a more unified and efficient way to integrate various information sources (image, video, text and LiDAR point cloud). Unlike traditional methods requiring separate models for different modalities, FMs provide a unified framework that simplifies multi-modal processing, leading to more comprehensive scene understanding and risk assessment \cite{cao2024maplm,wang2023accidentgpt}. This unified paradigm significantly reduces system complexity while enabling better cross-modal learning and feature transfer. The shared architectural framework facilitates more consistent interpretations across modalities and simplifies real-world deployment.

\subsubsection{Scene generation via FWMs}

Foundation World Models (FWMs), exemplified by systems like SORA, demonstrate sophisticated capabilities in simulating complex physical interactions and dynamic scenes while exhibiting a deep understanding of real-world principles \cite{cho2024sora,wang2024occsora}. These models showcase remarkable abilities in visual scene generation. A key advantage of FWMs in TSS is their ability to generate high-fidelity visual data for training perception models, particularly for rare but critical events that are challenging to capture in real-world datasets \cite{joshi2024synthetic}. Through controllable scene generation, these models can produce diverse visual scenarios spanning different lighting conditions, weather situations, and traffic configurations, which significantly enhance the robustness of perception systems. Furthermore, the synthetic data generated by FWMs holds promise for training visual detection systems targeting rare events such as traffic violations, accidents, and near-miss scenarios. By providing large-scale, diverse, and accurately annotated training data, these models help overcome the data scarcity challenge in developing reliable event detection systems. 

Moreover, FWMs can significantly enhance models' scene understanding and reasoning capabilities through their sophisticated simulation abilities \cite{chen2024towards}. By generating diverse sequences of traffic scenarios with explicit causal relationships, models can learn to better comprehend complex spatial-temporal interactions and identify critical risk factors \cite{guo2021computer,wen2021quantifying}. This systematic exposure to varied causal chains enables models to develop more nuanced understanding of traffic dynamics, leading to improved capabilities in both event detection and situation interpretation. Such enhanced understanding is particularly valuable for developing more intelligent surveillance systems that can anticipate potential risks rather than simply detecting events after occurrence \cite{noh2022novel}.

\subsubsection{Challenges and Solutions}

Despite their transformative potential, FMs face several critical challenges in TSS deployment that require systematic solutions:

\textbf{Computational overhead} represents a primary concern, as FMs typically demand substantial computational resources that may conflict with real-time processing requirements in TSS applications. Solutions include developing TSS-specific lightweight FMs through model compression and distillation techniques, implementing edge-optimized architectures, and adopting hierarchical deployment strategies where complex reasoning is performed offline or in cloud environments while lightweight models handle real-time tasks.

\textbf{Domain adaptation issues} present another significant challenge, as FMs trained on general datasets may not directly transfer to traffic-specific scenarios without substantial fine-tuning. Addressing this requires developing TSS-specific fine-tuning datasets, implementing progressive fine-tuning approaches that balance performance with resource constraints, and establishing transfer learning protocols that leverage traffic domain knowledge effectively.

\textbf{Interpretability and reliability concerns} are particularly critical in safety-sensitive TSS applications, where model decisions must be transparent and trustworthy. Mitigation strategies include implementing uncertainty quantification in FM outputs, developing ensemble approaches that combine FMs with domain-specific models, establishing robust evaluation frameworks with TSS-specific benchmarks, and incorporating human-in-the-loop validation for safety-critical applications to prevent potential hallucination risks.

\textbf{Synthetic-real domain gaps} specific to FWMs require careful attention to ensure that generated scenarios accurately reflect real-world traffic dynamics. Solutions include developing traffic-specific FWMs trained on comprehensive real TSS data, implementing domain adaptation techniques for synthetic data, and establishing validation frameworks that rigorously test the fidelity of synthetic scenarios against real-world observations. Addressing these limitations systematically will be crucial for realizing FMs' full potential in advancing TSS capabilities while maintaining the safety and reliability standards required in traffic surveillance applications.

\section{Conclusion}

This comprehensive review has systematically examined the current research, challenges, and future directions of vision technologies in TSS. Our analysis reveals that while significant progress has been made in both low-level and high-level perception tasks, five fundamental limitations persist: perceptual data degradation, data-driven learning constraints, semantic understanding gaps, sensing coverage limitations and computational resource demands. 

Current research efforts have explored approaches to address these challenges, though gaps remain: \textit{advanced perception enhancement} techniques (e.g., image enhancement, domain adaptation) show promise for challenging conditions but face deployment feasibility issues with high computational costs and struggle with robust generalization across diverse adverse conditions; \textit{efficient learning paradigms} (e.g., few-shot learning, self-supervised methods) offer pathways to reduce data dependency while requiring careful consideration of domain transfer and the synthetic-to-real gap; \textit{knowledge-enhanced understanding} approaches (e.g., spatiotemporal modeling, scene graph generation) provide frameworks for bridging semantic gaps yet struggle with modeling complex multi-agent interactions while ensuring causality and real-time performance; \textit{cooperative sensing frameworks} demonstrate potential for expanding coverage through multi-source fusion and multi-view collaboration but encounter challenges in data heterogeneity, temporal synchronization, and robust feature alignment across varied perspectives; and \textit{efficient computing frameworks} optimize resource utilization through lightweight model design, model compression, and distributed computing while addressing TSS-specific robustness requirements and communication overhead management. Moreover, the emergence of foundation models offers transformative potential in TSS, as they exhibit remarkable zero-shot learning abilities, strong generalization, and sophisticated reasoning capabilities across diverse tasks.

Looking forward, TSS development will likely focus on strategic integration of these complementary approaches to create more robust and deployment-ready systems, advancing knowledge-enhanced frameworks that capture complex traffic dynamics through scalable interaction models and temporal reasoning, developing adaptive cooperative sensing architectures that intelligently handle data quality variations and geometric complexities, and optimizing TSS-specific computing frameworks that dynamically balance performance with resource constraints through domain-aware optimization and enhanced edge intelligence. This evolution, combining traditional approaches with emerging technologies, will be crucial for advancing intelligent transportation infrastructure while addressing practical challenges in real-time performance, multi-modal data fusion, and deployment feasibility under diverse real-world conditions.

\begin{acks}
  This research is supported by the Major Research Plan of National Natural Science Foundation of China (92370203), the National Key Research and Development Program of China (2023YFE0106800) and the Science Fund for Distinguished Young Scholars of Jiangsu Province (BK20231531).
\end{acks}

\bibliographystyle{ACM-Reference-Format}
\bibliography{main}

% \appendix








% \end{spacing} 
\end{document}